%% file: main.tex
\documentclass[10pt,twocolumn,letterpaper]{article}

\usepackage{cvpr}              


\usepackage{algorithm}
\usepackage{algorithmic}
\usepackage{graphicx}
\usepackage{bm}
\usepackage{multirow}
\usepackage{amsmath}
\usepackage{amssymb}
\usepackage{pifont}
\usepackage{makecell}

\newsavebox\CBox

\definecolor{cvprblue}{rgb}{0.21,0.49,0.74}
\usepackage[pagebackref,breaklinks,colorlinks,allcolors=cvprblue]{hyperref}

\def\paperID{} 
\def\confName{CVPR}
\def\confYear{2025}

\title{From Histopathology Images to Cell Clouds: Learning Slide Representations with Hierarchical Cell Transformer}

\author{Zijiang Yang$^{1,2,}$\thanks{Equal Contribution} \thanks{This work was done when Zijiang Yang conducted an internship at DAMO Academy, Alibaba Group}, Zhongwei Qiu$^{2,3,6,*}$, Tiancheng Lin$^{2,3,7,*}$, Hanqing Chao$^{2,3,7,*}$,\\
Wanxing Chang$^{2,3}$,
Yelin Yang$^{4}$, Yunshuo Zhang$^{4}$, Wenpei Jiao$^{2,8}$, Yixuan Shen$^{5}$,\\
Wenbin Liu$^{5}$, Dongmei Fu$^{1}$, Dakai Jin$^{2}$, Ke Yan$^{2,3}$, Le Lu$^{2}$, Hui Jiang$^{4}$, Yun Bian$^{5}$\\
{\small $^{1}$University of Science and Technology Beijing, $^{2}$DAMO Academy, Alibaba Group, $^3$Hupan Lab}\\
{\small $^{4}$Department of Pathology, Changhai Hospital, $^{5}$Department of Radiology, Changhai Hospital,}\\
{\small $^{6}$Zhejiang University, $^{7}$Fudan University, $^{8}$Peking University}\\
{\tt\small \{qiuzhongwei.qzw, hanqing.chq\}@alibaba-inc.com}
}

\begin{document}
\maketitle
\input{sec/0_abstract}    
\input{sec/1_intro}
\input{sec/2_related_work}
\input{sec/3_dataset}
\input{sec/5_method}
\input{sec/6_exps}
\input{sec/7_conclusion}

\appendix
\section{Appendix}
\subsection{Pseudocode of HSP}
\label{sec:app_pseudo}

The pseudocode of Hierarchical Spatial Perception (HSP) is shown as Algorithm \ref{alg:HSP}.
HSP encodes input features with a fully connected layer and learns the cell spatial distribution at multiple scales via a hierarchical method.
Each level of HSP has a similar structure.
First, we generate anchors and split input points into multiple groups. Each group represents a local region of WSI.
Second, we filter points by applying the semantic-spatial aware filter and update point features to learn the cell spatial distribution within each group.
Third, we compute the higher-level features by the group-wise aggregation.
Finally, after repeating the above process $L$ times, the cell spatial distribution feature of the WSI is the maximum aggregation of the last-level features.

\subsection{Evaluation of the WSI-Cell5B}
\label{sec:app_cell5b}

\begin{figure}[t]
    \centering
    \includegraphics[width=0.99\linewidth]{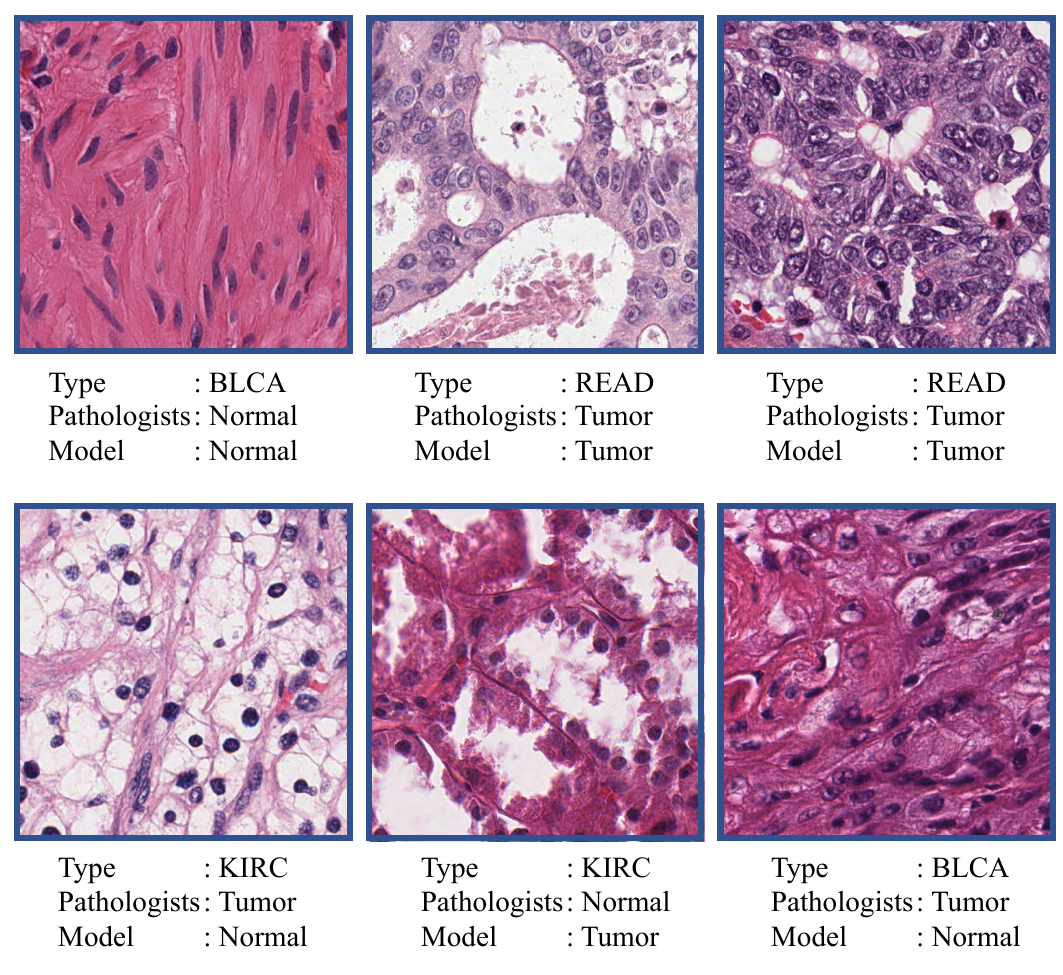}
    \caption{
    \textbf{Some cases of patches.}
    The first row shows cases where  WSI-Cell5B annotations are consistent with the diagnoses of pathologists. In contrast, the second row shows cases where WSI-Cell5B annotations differ from the diagnoses of pathologists.
    Pathologists and WSI-Cell5B have consistent judgments on over 92\% of the patches in the test set.
    }
    \label{fig:app_hard_case}
\end{figure}

We evaluate the annotation accuracy of WSI-Cell5B at the patch level.
Similarly to the Weakly Supervised Label Refinement (WSLR) proposed to refine annotations, pathologists are asked to label patches in the test set as normal, tumor, or indeterminate.
If a sample is deemed "indeterminate" by any pathologist, we mark it as an invalid sample.
We only use valid patches to evaluate the WSI-Cell5B.
The ground truth label of each patch is the voting of pathologists.
For WSI-Cell5B, if the proportion of cancer cells in a patch exceeds 25\%, the patch is labeled as tumor, otherwise, it is labeled as normal.
In our experiments, the accuracy of the annotations of WSI-Cell5B is 92.7\%.
Figure \ref{fig:app_hard_case} shows some cases of patches.
WSI-Cell5B annotations differ from diagnoses of pathologists on a few patches that are overly stained and have unclear tissues.
Experiments on survival prediction and cancer staging show that WSI-Cell5B is sufficient for analyzing cell spatial distributions and achieving state-of-the-art (SOTA) performance.


\begin{algorithm}[tb]
\caption{Hierarchical Spatial Perception (HSP)}
\label{alg:HSP}
\textsc{Iuput}: Coordinates of cells $C = \{c^{(i)}\}_{i=1}^{N_{total}}$, features of cells $F_{cell} = \{f_{cell}^{(i)}\}_{i=1}^{N_{total}}$.\\
\textsc{Parameter}: The number of group anchors $N_k$, filter threshold $\lambda_{sim}$, the basic number of points within each group $N_{basic}$, the number of perception levels $L$.
\begin{algorithmic}[1]
\STATE Encode features $F = \mathcal{F}_{fc}(F_{cell})$, where $\mathcal{F}_{fc}$ is a fully connected layer.
\FOR{level $l = 1, \cdots, L$}
\STATE Generate coordinate of group anchors $C_k$ by FPS.
\STATE Group points according to $C$ and $C_k$.
\FOR{group $k = 1, \cdots, N_k$}
\STATE Compute $S_{sim}$ according to Equation \ref{eq:compute_mask} of the main text.
\STATE Create $M = bool(S_{sim} > \lambda_{sim})$.
\STATE Update $F$ according to Equation \ref{eq:att} of the main text.
\STATE Compute spatial distribution of $k$-th group $f_{group}^{(k)}$ with average aggregation.
\ENDFOR
\STATE $C \leftarrow C_k$.
\STATE $F \leftarrow \{f_{group}^{(k)}\}_{k=1}^{N_k}$
\STATE $N_k \leftarrow N_k / N_{basic}$
\ENDFOR
\STATE The WSI cell spatial distribution feature $f_{cell}^{(WSI)} = \text{MaxAgg}(F)$.
\end{algorithmic}
\end{algorithm}

\begin{figure}[t]
    \centering
    \includegraphics[width=0.99\linewidth]{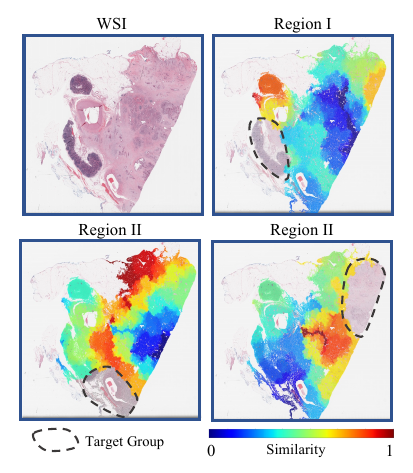}
    \caption{
    \textbf{Visualization of similarity.}
        CCFormer effectively learns the semantics of different regions.
    }
    \label{fig:att_score}
\end{figure}

\subsection{Visualization of Attention}
\label{sec:vis_attention}

To illustrate how CCFormer understands relationships across different regions, we visualize the similarity among groups at the last level.
Specifically, we select a group and visualize the similarity between this group and others.
In addition, the attention scores are mapped back to the input cell cloud.
As shown in Figure \ref{fig:att_score}, CCFormer comprehends the semantic relationships among regions.

\subsection{Clinical Interpretation of the Main Results}
\label{sec:cli_analysis_main}

WSI-Cell5B is designed to describe as many types of cancer as possible. 
Therefore, WSI-Cell5B categorizes cells into three basic types: neoplastic, inflammatory, and other.
Experiments on survival prediction and cancer staging validate the rationality of this design, and methods based on cell cloud achieve SOTA methods on most cancer types.

This design also limits the performance of methods based on cell cloud in some cancers, which require fine-grained cell classification.
For KIRC, the nuclear grade is significantly associated with patient survival risk~\cite{frank2002outcome, sorbellini2005postoperative}.
For BLCA, depth of invasion is a critical indicator in cancer staging~\cite{dyrskjot2023bladder}.
The stage of BLCA WSIs can be effectively judged by further categorizing neoplastic cells based on whether there is muscle invasion.
Due to the absence of these specific cell types, methods based on cell cloud performed inferiorly compared to MIL-based methods in the main results.
In contrast, for highly heterogeneous cancer types, such as PAAD~\cite{samuel2012molecular}, methods based on cell cloud can learn this heterogeneous cell spatial distribution and apply it to downstream tasks, thereby outperforming MIL-based methods.
Therefore, further subclassification of cells represents an effective strategy to enhance the performance of methods based on cell cloud in downstream tasks.
We will explore it in our future work.




\subsection{Implementation}
\label{sec:app_impl}

\subsubsection{WSI-Cell5B}

\noindent \textbf{Data Annotation.} 
We pre-process and detect cells with the same workflow for each WSI:

\begin{itemize}

\item \textbf{Region of Interest.}
We employ CLAM~\cite{lu2021data} to extract Regions of Interest (ROI) in WSIs to reduce the number of pixels that need to be processed for cell detection and classification.
WSIs are fixed at 40x magnification to ensure that each cell has sufficient detail.
Due to the limitations of memory, the ROI of each WSI is divided into patches of 512$\times$512 pixels for further processing.

\item \textbf{Cell Detection and Classification.}
As the data distribution of the PanNuke~\cite{gamper2020pannuke} partially overlaps with TCGA, we pre-train DPA-P2PNet~\cite{shui2024dpa} on it to improve the accuracy of cell detection and classification.
In this work, we focus on analyzing the spatial relationship among neoplastic cells, inflammatory cells, and other cells.
Therefore, we integrate the cell types of PanNuke in the pre-training. Specifically, cells with the type of connective, dead, and epithelial in PanNuke are labeled as others.

\item \textbf{Merging Patches.}
We further merge the prediction on patches and generate results for each WSI.
Since the same cell might be split into multiple patches and predicted repeatedly by the model, we merge cells that are close to the patch boundaries.
Specifically, the image resolution of WSI at 40x magnification is about 0.25 $\mu$m per pixel and the cell diameter is approximately 10 $\mu$m.
Therefore, we select cells that are less than 24 pixels away from the patch boundaries and merge cells of the same type that are less than 12 pixels apart.

\end{itemize}

\noindent \textbf{Post-Process.}
The WSI-Cell5B dataset has been stored in various formats for convenience, including images for preview, GeoJson files for visualization, and H5 files for fast loading.
In addition, following the method to pre-process large-scale point cloud datasets~\cite{dai2017scannet}, we also provide point clouds after grid sampling to improve analysis efficiency.
Specifically, the grid size is 256 pixels at 40x magnification.
Unless otherwise specified, we default to using the downsampled cell clouds for experiments.

\subsubsection{Clinical Analysis}

We introduce a hyper-parameter vector $\bm{\alpha}$ to control the weight of each component in the Cell Proportion Score (CPS) and Multi-scale Cell Proportion Score (MCPS).
We focus on the ratio of neoplastic cells to inflammatory cells on PAAD while giving greater weight to the proportion of inflammatory cells on HNSC. 
Thus, for HNSC, KIRC, and PAAD, $\bm{\alpha}$ is set to $[0.33, 0.33, 0.33]^T$, $[0.25, 0.50, 0.25]^T$, and $[0.0, 0.0, 1.0]$, respectively.
Additionally, on HNSC, the proportion of inflammatory cells, the proportion of neoplastic cells, and the ratio of neoplastic cells to inflammatory cells are given equal weight.
In practice, the number of bounding boxes in MCPS is set to 20. The size and coordinates of each box are computed by randomly sampling a ratio between 0.6 and 1.0 relative to the original cell cloud.

\begin{table}[!ht]
    \centering
    \small
    \renewcommand{\tabcolsep}{6pt}
    \renewcommand{\arraystretch}{1.3}

    \resizebox{\linewidth}{!}{

    \begin{tabular}{lc|cc}

    \toprule

    & Method & BLCA & \makecell[c]{COAD \\ READ} \\

    \midrule

    \parbox[t]{2mm}{\multirow{4}{*}{\rotatebox[origin=c]{90}{{\textbf{Patch feature MIL}}}}} 
    & MeanPool (Patch) & 0.514 $\pm$ 0.053 & 0.305 $\pm$ 0.067 \\
    & MaxPool (Patch) & 0.388 $\pm$ 0.080 & 0.262 $\pm$ 0.059 \\
    & ABMIL~\cite{ilse2018attention} & 0.536 $\pm$ 0.105 & 0.369 $\pm$ 0.052 \\
    & TransMIL~\cite{shao2021transmil} & \underline{0.540} $\pm$ 0.052 & 0.298 $\pm$ 0.051 \\
    \midrule
    \parbox[t]{2mm}{\multirow{2}{*}{\rotatebox[origin=c]{90}{{\textbf{Graph}}}}} 
    & Patch-GCN~\cite{chen2021whole} & 0.466 $\pm$ 0.064 & 0.229 $\pm$ 0.050\\
    & WiKG~\cite{li2024dynamic} & 0.518 $\pm$ 0.027 & 0.345 $\pm$ 0.069 \\
    \midrule
    \parbox[t]{2mm}{\multirow{5}{*}{\rotatebox[origin=c]{90}{{\textbf{Point Cloud}}}}} 
    & MeanPool (Cell) & 0.389 $\pm$ 0.029 & 0.216 $\pm$ 0.039 \\
    & MaxPool (Cell) & 0.246 $\pm$ 0.059 & 0.229 $\pm$ 0.055 \\
    & PointNet~\cite{qi2017pointnet} & 0.434 $\pm$ 0.032 & 0.310 $\pm$ 0.059 \\
    & PointNet++~\cite{qi2017pointnetpp} & 0.388 $\pm$ 0.042 & 0.262 $\pm$ 0.022 \\
    & PTv3~\cite{wu2024point} & 0.369 $\pm$ 0.090 & 0.262  $\pm$ 0.055 \\
    \midrule
    & \textbf{CCFormer (ours)} & 0.487 $\pm$ 0.025 & \textbf{0.396} $\pm$ 0.015 \\
    & \makecell[c]{\textbf{CCFormer (ours)} \\ \textbf{+ MeanPool (Patch)}} & \textbf{0.560} $\pm$ 0.040 & \underline{0.370} $\pm$ 0.031 \\

    \bottomrule
    
    \end{tabular}
    }

    \caption{Comparison of cancer staging with SOTA methods on BLCA and COADREAD in Macro-F1 ($\uparrow$).
    The combination of CCFormer and the global mean pooling outperforms baselines on both cancers.
    }
    \label{tb:staging_tb}
    
\end{table}

\subsubsection{CCFormer}

The scale factor $\lambda_{r} = 4$ and the discrete number $N_d = 3$ in NIE.
For HSP, we set the number of perception levels $L=3$, the initial number of group anchors $N_k=2048$, and the basic number of points within each group $N_{basic}=16$. In each level, the input features are updated twice based on Equation \ref{eq:att} of the main text, and the input dimension is expanded to twice its original size in the group-wise aggregation.
The semantic-spatial aware filter threshold $\lambda_{sim}$ is set to 0.5 by default.

To combine CCFormer and MeanPool (Patch), the global average feature of patches is passed through two fully connected layers to get the WSI appearance feature $f_{app}$, which has the same dimension as the WSI cell spatial distribution feature.
Therefore, the WSI feature $f_{wsi} = f_{cell}^{(WSI)} + \beta f_{app}$, where $\beta$ is the weight of the WSI appearance feature.

\begin{table}[t]
    \centering
    \small
    \renewcommand{\tabcolsep}{3pt}
    \renewcommand{\arraystretch}{1.3}

    \resizebox{\linewidth}{!}{

    \begin{tabular}{cc}

    \toprule

    Abbreviation & Full Name \\

    \midrule

    BLCA & Bladder Urothelial Carcinoma\\
    BRCA & Breast Invasive Carcinoma\\
    COAD & Colon Adenocarcinoma\\
    HNSC & Head and Neck Squamous Cell Carcinoma\\
    KIRC & Kidney Renal Clear Cell Carcinoma\\
    LUAD & Lung Adenocarcinoma\\
    LUSC & Lung Squamous Cell Carcinoma\\
    PAAD & Pancreatic Adenocarcinoma\\
    READ & Rectum Adenocarcinoma\\
    STAD & Stomach Adenocarcinoma\\
    UCEC & Uterine Corpus Endometrial Carcinoma\\

    \bottomrule
    
    \end{tabular}
    }

    \caption{Cancer abbreviation and full name cross-reference table.
    }
    \label{tb:cancer_name}
    
\end{table}

\subsubsection{Training}

We follow the splitting method of SurvPath~\cite{jaume2023modeling} that WSIs are split into 5 folds according to the sample sit. Negative log-likelihood survival loss and cross-entropy loss are employed for training models on survival prediction and cancer staging, respectively.

Baselines~\cite{ilse2018attention, shao2021transmil, chen2021whole, li2024dynamic, qi2017pointnet, qi2017pointnetpp, wu2024point} are implemented with their released codes.
MIL-based methods are optimized using RAdam~\cite{liu2019variance}, a batch size of 1, a learning rate of $2\times10^{-4}$, $1\times 10^{-3}$ weight decay, and epochs of 20.
Patch-GCN~\cite{chen2021whole} and WiKG~\cite{li2024dynamic} are optimized using their default hyper-parameters.
Point cloud methods are optimized using Adam~\cite{kingma2014adam}, a batch size of 8, a learning rate of $1 \times 10^{-3}$, cosine annealing learning rate decay to $1 \times 10^{-4}$, and epochs of 150.
For CCFormer, we adjust the learning rate, the semantic-spatial aware filter threshold $\lambda_{sim}$, and the dropout ratio for each cancer based on the same training parameters as the point cloud methods.
Due to the significant difference in convergence speed between cell cloud methods and MIL-based methods, the combination of CCFormer and MeanPool (Patch) fails to model the cell spatial distribution if model parameters are randomly initialized.
Therefore, the combination model loads the pre-trained CCFormer and freezes it during training. Only the two fully connected layers added for MeanPool (Patch) are optimized using Adam, a batch size of 8, a learning rate of $5\times 10^{-4}$, cosine annealing learning rate decay to $1 \times 10^{-4}$, and epochs of 10.



\subsection{Detailed Results}
\label{sec:app_res}

Table \ref{tb:staging_tb} reports the detailed cancer staging results of Figure \ref{fig:staging} in the main text. CCFormer outperforms baselines.

\subsection{Symbol Explanation}

Table \ref{tb:cancer_name} summarizes the cross-reference of cancer abbreviation and full name.

\clearpage
{
    \small
    \bibliographystyle{ieeenat_fullname}
    \bibliography{main}
}


\end{document}

%% file: sec/0_abstract.tex
\begin{abstract}
It is clinically crucial and potentially beneficial to analyze and directly model the spatial distributions of cells in histopathology whole slide images (WSI).
However, most existing WSI datasets lack cell-level annotations, owing to the extremely high cost over gigapixel images.
Thus, it remains an open question whether deep learning models can directly and effectively analyze WSIs from the semantic aspect of cell distributions.
In this work, we construct a large-scale WSI dataset (\textbf{WSI-Cell5B}) with more than 5 billion cell-level annotations and a novel hierarchical Cell Cloud Transformer (\textbf{CCFormer}) to tackle these challenges.
WSI-Cell5B is based on 6,998 WSIs of 11 cancers from {The Cancer Genome Atlas Program}, and all WSIs are annotated per cell with coordinates and types.
To the best of our knowledge, WSI-Cell5B is the first WSI-level large-scale dataset integrating cell-level annotations.
Besides, CCFormer formulates the collection of cells in each WSI as a cell cloud to model cell spatial distribution.
In CCFormer, Neighboring Information Embedding is proposed to characterize the distribution of cells within the neighbor of cells, and a Hierarchical Spatial Perception module is proposed to learn the spatial relationship among cells in a bottom-up manner.
Clinical analysis indicates that WSI-Cell5B can be used to design clinical evaluation metrics based on counting cells that effectively assess patients' survival risk.
Extensive experiments on survival prediction and cancer staging show that learning from cell spatial distribution alone can already achieve state-of-the-art performance, i.e., CCFormer evidently outperforms other competing methods.
\end{abstract}

%% file: sec/1_intro.tex
\section{Introduction}\label{sec:intro}

Analyzing histopathology whole slide images (WSIs) presents a significant challenge in computational pathology. It requires managing gigapixel images while capturing the features and distributions of tissues and cells.
Significant progress in WSI analysis and related downstream tasks has been achieved by training models on high-quality WSI datasets, such as those from the Cancer Genome Atlas Program (TCGA)~\cite{liu2018integrated}. These advancements include tasks like survival prediction~\cite{chen2021whole, shao2024tumor}, cancer staging~\cite{li2024dynamic}, cancer sub-typing~\cite{song2024morphological}, and gene mutation prediction~\cite{xu2024whole}.

\begin{figure}[t]
    \centering
    \includegraphics[width=0.99\linewidth]{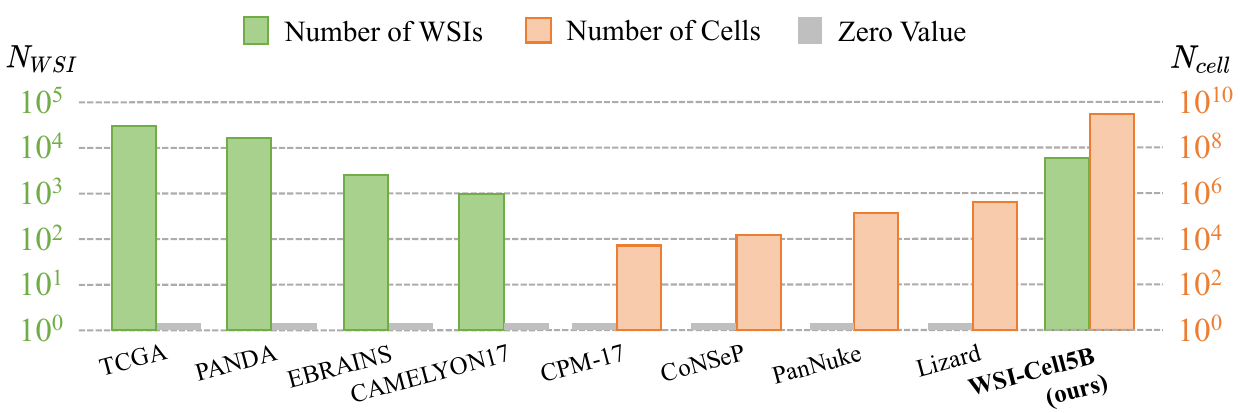}
    \caption{
    \textbf{Comparison of histopathology datasets on the number of cells and WSIs.} Our proposed WSI-Cell5B is the first WSI-level large-scale dataset integrating cell-level annotations, while existing datasets lack either cell-level annotations or WSIs for clinical endpoints.}
    \label{fig:first_fig}
\end{figure}

Existing methods~\cite{ilse2018attention, chen2021whole,chen2022scaling} typically analyze  WSIs via conventional image perception frameworks, where the image representation is the cornerstone of downstream tasks.
Thus, numerous histological foundation models~\cite{ xu2024whole,chief,chen2024towards,lu2024avisionlanguage,pathchat} have been proposed, pre-trained on large-scale datasets for general-purpose representations.
Unlike natural images, the analysis of \textit{cell spatial distribution} within WSIs has been verified as clinically important, associated with the molecular profile~\cite{saltz2018spatial}, tumor progression~\cite{corredor2019spatial}, prognostic biomarkers~\cite{page2023spatial}, $etc$. 
Heavy reliance on the foundation models, combined with an oversight of cell spatial distribution, results in high computational costs and suboptimal performance.
Fundamentally, this limitation stems from the absence of cell-level annotations in existing WSI datasets~\cite{liu2018integrated, bandi2018detection, bulten2022artificial, roetzer2022digital} due to the extremely high cost~\footnote{Over 760,000 cells on a WSI of 50,000$\times$50,000 pixels.}.
As shown in Figure \ref{fig:first_fig}, existing  histopathology datasets are lack of either cell-level annotations ($e.g.$, TCGA) or clinical endpoints~\cite{vu2019methods, gamper2020pannuke, graham2021lizard, graham2019hover}.
Therefore, analyzing WSIs by modeling the cell spatial distribution remains an open problem.

We argue the collection of cells within a WSI can be regarded as a specific form of the point set, termed cell cloud, thus we can learn slide representations by cell cloud modeling.
To this end, we first provide WSI-Cell5B, a large-scale WSI dataset consisting of 6,998 WSIs of 11 cancers and over 5 billion cell-level annotations.
We further propose a novel weakly supervised label refinement method, based on foundation models, to reduce the cost of cell-level annotations.
Second, we notice that cell clouds exhibit a significant hierarchical structure: local cell clusters ($.e.g$, tumor cellularity), larger cell spatial distribution structures ($e.g.$, blood vessels) and the tissue microenvironment at the WSI level that can reflect clinical indicators such as cancer stage and patient survival risk.
This motivates us to propose a novel Hierarchical cell Transformer, termed CCFormer, to model the cell clouds.
CCFormer consists of two key modules: \textbf{Neighboring Information Embedding (NIE)} and \textbf{Hierarchical Spatial Perception (HSP)}.
NIE describes the neighborhood cell distribution pattern of cells at the cell level by evaluating the statistical characteristics of each type of cell within the cell neighborhood.
HSP further progressively perceives and aggregates cell spatial distribution information hierarchically.
The clinical analysis on WSI-Cell5B indicates that the survival risk of patients can be effectively decided by evaluating the proportions of various cell types, which is difficult to obtain based solely on WSI.
Extensive experiments on survival prediction and cancer staging show that analyzing WSIs via cell clouds is a highly competitive framework, and CCFormer outperforms other methods.
Our contributions can be summarized:
\begin{itemize}

    \item 
    We propose WSI-Cell5B, the first large-scale dataset to integrate cell-level annotations with WSIs, comprising 6,998 WSIs of 11 cancers and more than 5 billion cell-level annotations. 
    WSI-Cell5B can be used to analyze the cell spatial distribution of entire WSIs comprehensively.
    To reduce the cost of cell-level annotations, we propose a weakly supervised label refinement method based on foundation models.
    \item 
    We regard the collection of cells within WSIs as cell clouds and propose a novel hierarchical Cell Cloud Transformer, termed CCFormer, to model cell clouds.
    CCFormer introduces a novel Neighboring Information Embedding (NIE) to embed the neighborhood cell distribution at the cell level and a novel Hierarchical Spatial Perception (HSP) to model cell spatial distribution information in a bottom-up manner.

    \item The clinical analysis on WSI-Cell5B validates that WSI-Cell5B can be directly used to construct effective clinical indicators. In addition, extensive experiments verify the effectiveness of the cell cloud framework and CCFormer.

\begin{figure*}[t]
    \centering
    \includegraphics[width=0.95\linewidth]{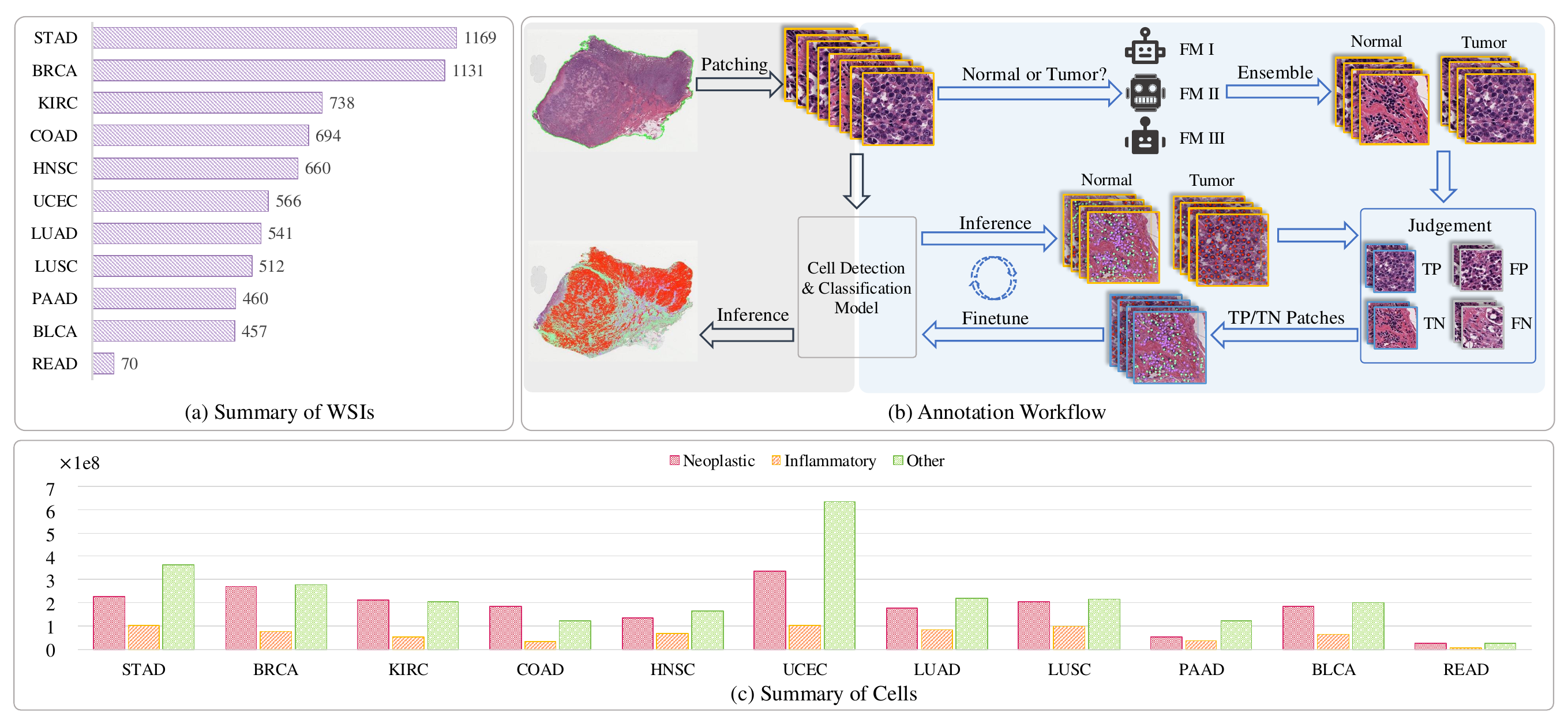}
    \caption{
    \textbf{Overview of WSI-Cell5B. }(a) Summary of WSIs within WSI-Cell5B. WSI-Cell5B includes 6,998 WSIs on 11 types of cancers across 10 organs.
    (b) The annotation workflow. A novel Weakly Supervised Label Refinement based on foundation models is proposed to reduce the costs of label refinement.
    (c) Summary of cells. WSI-Cell5B consists of over 5 billion cell-level annotations of three cell types. 
    }
    \label{fig:wsi-cell5b}
\end{figure*}
    
\end{itemize}

%% file: sec/2_related_work.tex
\section{Related Work}
\label{sec:related}


\noindent \textbf{Cell Detection and Classification Datasets.}
Due to the excessively high cost of cell-level annotation, current cell detection and segmentation datasets~\cite{kumar2017dataset, vu2019methods, gamper2020pannuke, graham2021lizard, graham2019hover} are composed of patches selected from WSIs.
For each patch, these datasets provide cell-level annotations, including the type, coordinates, and segmentation of cells and nuclei.
Specifically, PanNuke~\cite{gamper2020pannuke} consists of over 200,000 labeled nuclei and 19 different tissue types.
Lizard~\cite{graham2021lizard} consists of over 400,000 labeled nuclei of colon tissue.
However, as patches are independent, these datasets can not be used for analyzing cell spatial distribution of WSIs, limiting the application on downstream tasks, such as survival prediction, cancer staging, cancer sub-typing, and so on.
To build a high-accuracy cell annotation dataset across the entire WSI, reducing the cost of cell-level annotation is a critical issue.
In this paper, we propose a weakly supervised label refinement based on foundation models to address this issue.


\noindent \textbf{WSI Datasets.}
High-quality WSI datasets~\cite{liu2018integrated, bandi2018detection, bulten2022artificial, roetzer2022digital} consist of WSIs with detailed clinical information.
CAMELYON17~\cite{bandi2018detection} consists of over 1,000 WSIs for the detection and classification of breast cancer metastases.
PANDA~\cite{bulten2022artificial} consists of over 12,000 WSIs for Gleason grading of prostate biopsies.
EBRAINS~\cite{roetzer2022digital} consists of over 3,000 WSIs for brain tumor sub-typing.
TCGA~\cite{liu2018integrated} consists of over 40,000 WSIs from 28 organs.
These datasets have significantly facilitated the development of WSI analysis.
However, due to a lack of cell-level annotations, the cell spatial distribution of each WSI can not be analyzed based on these datasets.
In this paper, we propose WSI-Cell5B to track this issue.

\noindent \textbf{Patch-Level Methods in Histopathology.}
Patch-level methods~\cite{ilse2018attention, shao2021transmil, chen2021whole, li2024dynamic, shao2024tumor, chan2023histopathology, lin2023interventional,qiu2024end} divide WSIs into patches and employs pre-trained models~\cite{he2016deep, lu2024visual, chen2024towards, lin2023interventional} to extract patch features for downstream tasks.
Since WSIs are typically giga-pixel images, most existing methods~\cite{ilse2018attention, shao2021transmil} are designed with Multi-Instance Learning (MIL), where WSIs are formulated as a bag of sampled patch features.
Although MIL-based method can effectively analyze WSIs, these method only focus on sampled regions of interest, limiting to learning the spatial and semantic relationship of patches across the WSI.
To track this issue, graph of patches has been introduced into WSI analysis~\cite{chen2021whole, li2024dynamic, shao2024tumor, chan2023histopathology}.
Patch-GCN~\cite{chen2021whole} introduces patch-based graph convolutional networks to model the relationship among patches.
HEAT~\cite{chan2023histopathology} classifies patches and learns the relationship among patches via heterogeneous graph.
WiKG~\cite{li2024dynamic} introduces dynamic graph representation to update features of nodes and edges.
TMEGL~\cite{shao2024tumor} proposes a gated graph attention network to learn the micro-environment of patches.
Although graph-based methods can describe the relationships among patches, they are limited at the patch-level and unable to model the cell spatial distribution.
In addition, Ceograph~\cite{wang2023deep} propose to analyze cell spatial organization with graph. However, Ceograph focus on learning cell relationship within each patch and can not percept the cell distribution across the WSI.
In this paper, we formulate WSIs as cell clouds and propose to model cell spatial distribution across the entire WSI.

\noindent \textbf{Point Set Learning.} Point set learning aims to understand the spatial relationships between points in point sets, also known as point clouds.
Recently, deep learning approaches for learning point clouds have been rapidly developed and can be categorized into projection-based~\cite{qiu2022ivt, lang2019pointpillars}, voxel-based~\cite{maturana2015voxnet, choy20194d, graham20183d, chen2023largekernel3d}, point-based~\cite{qi2017pointnet, qi2017pointnetpp, ma2022rethinking, zhao2019pointweb,qiu2023learning, wu2022point}, and serialized methods~\cite{wu2024point, wang2023octformer, liang2024pointmamba}.
Since projection-based and voxel-based methods are typically designed for 3D point clouds, these methods are difficult to apply to 2D cell clouds.
While point-based methods can be easily extended to 2D cell clouds, existing methods are not suitable for describing the unique hierarchical spatial relationships among cells.
In addition, serialized methods organize points into sequences based on predefined patterns and lack flexibility in handling the varied hierarchical structures of cell clouds.
In this paper, we propose CCFormer, which progressively learns the relationships among cells hierarchically.

%% file: sec/3_dataset.tex
\section{The WSI-Cell5B Dataset}
\label{sec:dataset}

We collect H\&E-stained WSIs from TCGA~\cite{grossman2016toward} to build the WSI-Cell5B dataset.
As shown in Figure \ref{fig:wsi-cell5b} (a), WSI-Cell5B includes 6,998 WSIs on 11 types of cancers across 10 organs.
The statistics of cells are shown in Figure \ref{fig:wsi-cell5b} (c).
More than 5.2 billion cells (2.0 billion neoplastic cells, 0.7 billion inflammatory cells, and 2.5 billion others) have been identified. 
To improve the accuracy of annotations and reduce costs, We carefully design an annotation workflow as shown in Figure \ref{fig:wsi-cell5b} (b).
We perform preliminary annotations on WSIs with DPA-P2PNet~\cite{shui2024dpa} pre-trained on PanNuke~\cite{gamper2020pannuke}.
However, due to the differences in data distribution between PanNuke and TCGA, there are numerous errors in the preliminary annotations.
Moreover, due to the large scale of the cell-level annotations, manual label refinement would result in substantial costs.
Therefore, we propose a Weakly Supervised Label Refinement (WSLR) method based on foundation models to minimize human involvement and reduce the costs of label refinement.

\subsection{Weakly Supervised Label Refinement}

Label refinement at the cell level incurs significant manual costs. 
In contrast, annotating at the patch level is less expensive.
Therefore, WSLR utilizes samples with credible patch-level labels to fine-tune the pre-trained cell detection and classification models.
Specifically, WSLR involves a two-step cycle: 1) screening credible samples from the preliminary data annotations, and 2) fine-tuning.
With WSLR, the pre-trained model can further learn the characteristics of images from the WSI-Cell5B dataset, thus improving the accuracy of cell-level annotations.

In the first step, foundation models~\cite{lu2024avisionlanguage,ikezogwo2024quilt,sun2023pathasst} can determine whether patches are predominantly composed of cancer cells, based on appropriate prompts. 
Therefore, WSLR introduces judgments based on foundation models to validate the outcomes of cell detection and classification.
Given a patch, if the judgments based on foundation models and cell assessments are consistent, the patch is designated as a credible patch.
Specifically, the judgment based on foundation models is generated through a voting process among multiple models, whereas the cell-based assessment is achieved by quantifying the proportion of cancer cells.






\subsection{Clinical Analysis}

\begin{figure}[t]
    \centering
    \includegraphics[width=0.99\linewidth]{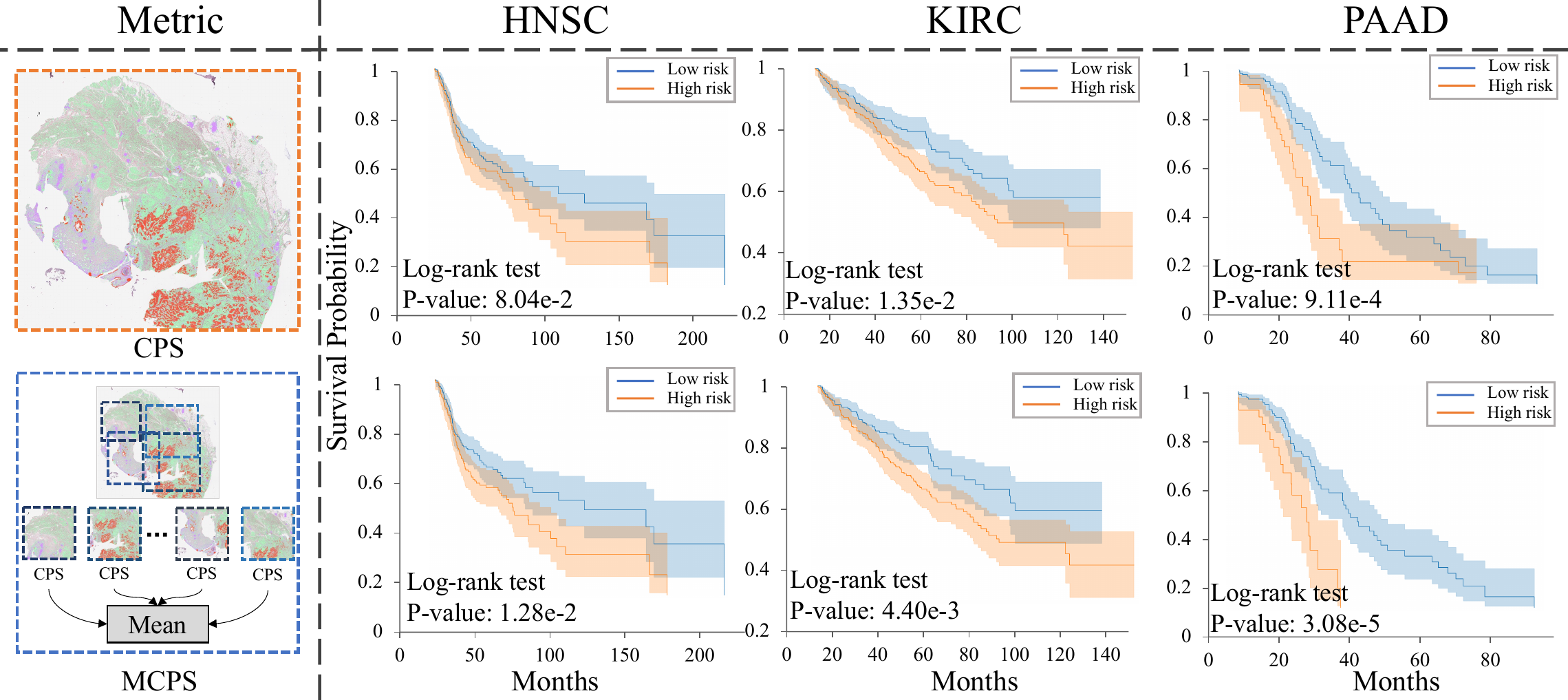}
    \caption{
    Kaplan-Meier analyses on HNSC, KIRC and PAAD.
    }
    \label{fig:clinical_analysis}
\end{figure}

Cell-level annotations and statistics hold significant clinical importance~\cite{page2023spatial}. 
To illustrate it, we construct survival risk evaluation metrics based on the WSI-Cell5B and conduct Kaplan-Meier analyses on three types of cancer.
Based on clinical experience, the survival risk is highly influenced by the proportion and distribution of neoplastic and inflammatory cells~\cite{page2023spatial}.
Therefore, We construct two metrics as shown in the left of Figure \ref{fig:clinical_analysis}: Cell Proportion Score (CPS) and Multi-scale Cell Proportion Score (MCPS).
CPS considers only the proportions of various cell types within the WSI, whereas the MCPS further considers the distribution of cells within small regions.

CPS is defined as follows:
\begin{equation}
\small
\label{eq:cps}
    S_{CPS} = [\frac{N_{neo}}{N_{total}}, \frac{N_{inf}}{N_{total}}, \frac{N_{neo}}{N_{inf}}] \bm{\alpha},
\end{equation}
where $S_{CPS}$ denotes Cell Proportion Score, $\bm{\alpha}=[\alpha_1, \alpha_2, \alpha_3]^T$ is the weight vector, and $N_{total}$, $N_{neo}$, $N_{inf}$ are the number of cells, neoplastic cells, and inflammatory cells within a WSI, respectively.
For different types of cancer, $\bm{\alpha}$ is set based on the type of cancer to focus on different components.
Furthermore, MCPS introduces randomly sized and positioned boxes to perceive the distribution of cells across different scales within the WSI: $S_{MCPS} = \frac{1}{N_{box}}\sum_{i=1}^{N_{box}}S_{CPS}^{(i)}$,
where $S_{MCPS}$ denotes Multi-scale Cell Proportion Score, $N_{box}$ is the number of bounding boxes, and $S_{CPS}^{(i)}$ denotes computing Cell Proportion Score within the $i$-th box.


As shown in the right of Figure \ref{fig:clinical_analysis}, CPS correctly distinguishes between high-risk and low-risk patients on PAAD and KIRC. However, due to the lack of local perception, SPC fails to distinguish patients of HNSC.
In contrast, MCPS successfully stratifies patients across the three cancer types and achieves lower p-values than CPS.
Compared to CPS, MCPS reduces the p-value from 9.11e-4, 1.35e-2, and 8.04e-2 to 3.08e-5, 4.40e-3, and 1.28e-2 on PAAD, KIRC, and HNSC, respectively.

%% file: sec/5_method.tex
\section{Hierarchical Cell Cloud Transformer}\label{sec:method}

\begin{figure*}[t]
    \centering
    \includegraphics[width=0.99\linewidth]{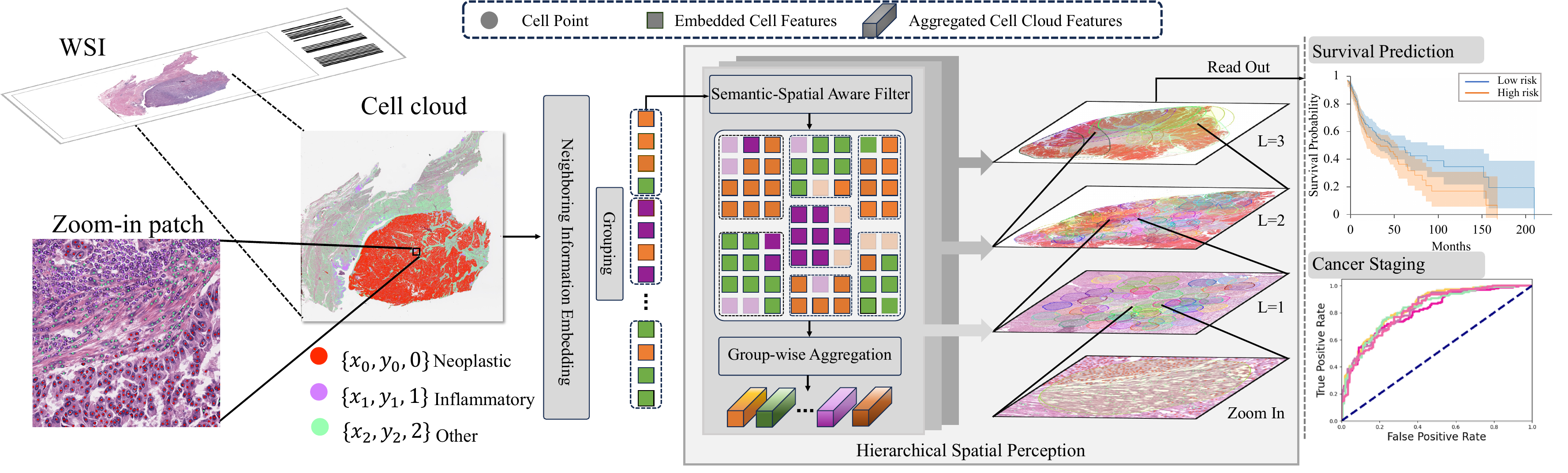}
    \caption{
    \textbf{The pipeline and illustration  of CCFormer.}
    Given the cell point (coordinate and type) within the cell cloud, Neighboring Information Embedding supplements the statistical characteristics of its neighboring cells.
    Hierarchical Spatial Perception further progressively perceives and aggregates cell spatial distribution information hierarchically.
    Finally, the feature of cell spatial distributions across the entire WSI is applied to clinical endpoints.
    }
    \label{fig:main_model_fig}
\end{figure*}

The framework of CCFormer is illustrated in Figure \ref{fig:main_model_fig}.
First, cell clouds are encoded with Neighboring Information Embedding (NIE) to supplement the neighborhood cell distribution.
The Hierarchical Spatial Perception (HSP) further learns and aggregates the cell spatial distributions hierarchically.
Finally, the feature of cell spatial distributions across the entire WSI is applied to clinical endpoints.


\subsection{Neighboring Information Embedding}

The neighborhood cell distribution patterns at the cell level are critical characteristics in distinguishing the cells with the same category.
For cells similarly labeled as cancer, whether they are surrounded by a large number of cancer cells or immune cells have completely different clinical significance~\cite{wang2023deep}.
Thus, we propose NIE to embed the spatial distribution information of neighboring cells.
Specifically, we propose the local and global density features to embed statistical information of neighboring cells.

For each WSI, the mean shortest distance among cells $d_{mean}$ across the dataset is used to adaptively set the largest local neighborhood radius $r_{max} = \lambda_r d_{mean}$, where $\lambda_r$ is the scale factor.
To obtain more precise local spatial information, we introduce discrete number $N_d$ to uniformly divide $r_{max}$ into multiple segments, thereby obtaining a series of radius $\textbf{r}=[r^{(1)}, r^{(2)}, \cdots r^{(N_d)}]^T$.
We denote the number of the $t$-th type of cell within the $j$-th radius of the $i$-th cell as $N^{(i, r^{(j)}, t)}, i\in\{1, 2, \cdots, C\}, j\in\{0, 1, 2, \cdots, N_d\}, t\in\{1, 2, \cdots, T\}$, where $C$ is the number of cells and $T$ is the number of cell types.
Specifically, $N^{(i, r^{(0)}, t)}=0$.
Thus, the local relative density feature is computed as follows:
\begin{equation}
\small
\label{eq:local_density_feature}
\begin{split}
    f_{ld}^{(i, r^{(j)}, t)} = \frac{N^{(i, r^{(j)}, t)} - N^{(i, r^{(j-1)}, t)}}{N^{(i, r^{(N_d)}, t)}},
\end{split}
\end{equation}
where $f_{ld}^{(*)}$ denotes the local density feature of the $t$-th type of cell within the $j$-th radius of the $i$-th cell. 
The local density feature vector of $i$-th cell $F_{ld}^{(i)}=[f_{ld}^{(i, r^{(1)}, 1)}, \cdots f_{ld}^{(i, r^{(N_d)}, T)}]^T$.
Equation \ref{eq:local_density_feature} measures the relative density of cells within multiple neighborhood radii, thereby quantifying the proximity between cells and their neighboring cells.
We further introduce the global density feature to quantify  the statistical distribution of cells across the entire cell cloud:
\begin{equation}
\small
\label{eq:global_density_feature}
\begin{split}
    f_{gd}^{(i, r^{(j)}, t)} = \frac{N^{(i, r^{(j)}, t)} - N^{(i, r^{(j-1)}, t)}}{\max_{i\in\{1,2,\cdots,C\}}(N^{(i, r^{(N_d)}, t))})},
\end{split}
\end{equation}
where $f_{gd}^{(*)}$ denotes the global density feature of the $t$-th type of cell within the $j$-th radius of the $i$-th cell.
The global density feature vector of $i$-th cell $F_{gd}^{(i)}=[f_{gd}^{(i, r^{(1)}, 1)}, \cdots f_{gd}^{(i, r^{(N_d)}, T)}]^T$.
In CCFormer, the embedding feature of each cell $F_{cell}$ is the concatenation of $F_{ld}$, $F_{gd}$, and the one-hot encoding of cell type.

\begin{figure}[t]
    \centering
    \includegraphics[width=0.95\linewidth]{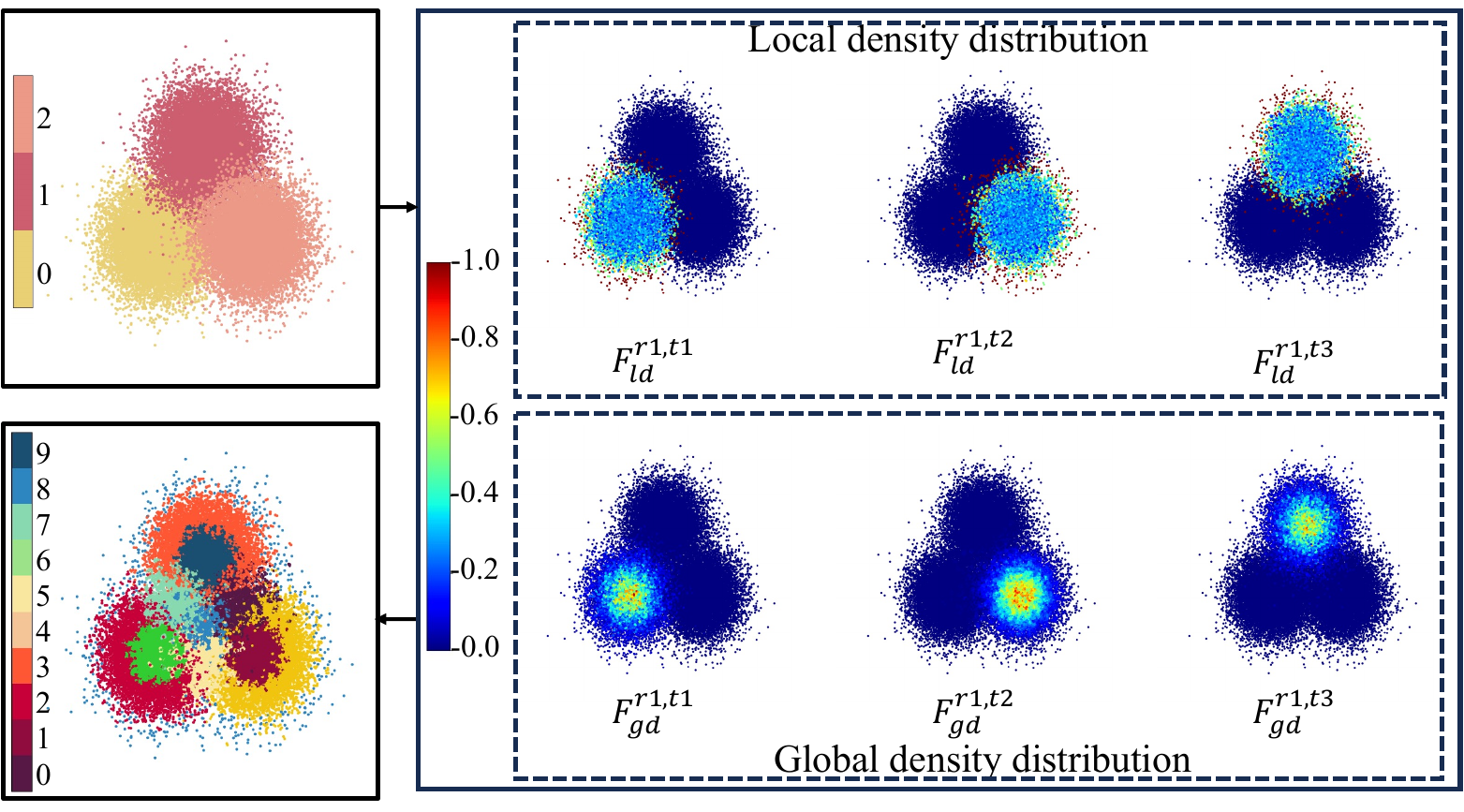}
    \caption{
    \textbf{Toy example of NIE.} 
    A toy point set containing three categories is generated. After extracting features via NIE, we performed K-Means clustering. The results indicate that features derived from NIE can effectively differentiate points at different locations (boundaries, core regions, and outliers).
    }
    \label{fig:feature_toy_example}
\end{figure}

We generate toy point sets with Gaussian distribution to illustrate NIE.
As shown in Figure \ref{fig:feature_toy_example}, NIE not only correctly distinguishes points of different categories but also further differentiates points of the same type located in different neighborhood patterns.
In Section \ref{sec:exps}, we further validate that the NIE can also effectively describe the neighborhood cell distribution on real cell clouds.



\subsection{Hierarchical Spatial Perception}

In WSI, the spatial distribution of cells is hierarchical.
The entire WSI is composed of multiple important regions, each of which consists of smaller clusters of cells.
We propose HSP to learn this hierarchical structure of cell clouds.
Specifically, HSP groups the cells to divide the WSI into a collection of sub-regions.
For each group, we conduct intra-group information interaction to learn the cell spatial distribution of the local region.
By aggregating features for each group and repeating the above process at higher levels, HSP hierarchical models cell clouds.

Given the coordinates $C = \{c^{(i)}\}_{i=1}^{N_{total}}$ and features $F_{cell} = \{f_{cell}^{(i)}\}_{i=1}^{N_{total}}$ of a collection of cells, group anchors $K = \{k^{(i)}\}_{i=1}^{N_k}$ are generated by Farthest Point Sampling (FPS)~\cite{qi2017pointnetpp}, where $c^{(i)}$ and $f_{cell}^{(i)}$ are the coordinate and feature of the $i$-th cell, $N_k$ is the number of group anchors, and $k^{(i)}$ is the $i$-th anchor.
To preserve the spatial distribution of cells as much as possible, we incorporate cell category information into the anchor generating.
For any group anchor $k^{(i)}$, $2N_{total}/N_k$ nearest neighbors are assigned to it and marked as a group.

Cells assigned to the same group are spatially proximate, thus forming a local region.
Consequently, we further update the cell features within each group and aggregate them to obtain features that describe the local cell spatial distribution.
However, multiple clusters of spatially adjacent cells may be assigned to the same group.
We introduce a semantic-spatial aware filter to address this issue and provide a detailed visualization and analysis in Section \ref{sec:exps}.

\begin{table*}[t]
    \centering
    \small
    \renewcommand{\tabcolsep}{7pt}
    \renewcommand{\arraystretch}{2}

    \resizebox{\textwidth}{!}{

    \begin{tabular}{lcc|cccccccccc}

    \toprule

    & Method & \makecell[r]{Params.} & BLCA & BRCA & \makecell[r]{COAD \\ READ}  & HNSC & KIRC & LUAD & LUSC & PAAD & STAD & UCEC \\

    \midrule

    \parbox[t]{2mm}{\multirow{4}{*}{\rotatebox[origin=c]{90}{{\textbf{Patch feature MIL}}}}} 
    & \makecell[c]{MeanPool (Patch)} & 4.1K & \makecell[r]{0.610 \\ $\pm$ 0.024} & \makecell[r]{0.643 \\ $\pm$ 0.031} & \makecell[r]{0.629 \\ $\pm$ 0.186} & \makecell[r]{0.572 \\ $\pm$ 0.037} & \makecell[r]{0.676 \\ $\pm$ 0.051} & \makecell[r]{0.571 \\ $\pm$ 0.057} & \makecell[r]{0.526 \\ $\pm$ 0.051} & \makecell[r]{0.672 \\ $\pm$ 0.097} & \makecell[r]{0.579 \\ $\pm$ 0.085} & \makecell[r]{0.727 \\ $\pm$ 0.041} \\

    & \makecell[c]{MaxPool (Patch)} & 4.1K & \makecell[r]{0.510 \\ $\pm$ 0.038} & \makecell[r]{0.589 \\ $\pm$ 0.063} & \makecell[r]{0.585 \\ $\pm$ 0.078} & \makecell[r]{0.561 \\ $\pm$ 0.068} & \makecell[r]{0.590 \\ $\pm$ 0.076} & \makecell[r]{0.480 \\ $\pm$ 0.037} & \makecell[r]{0.493 \\ $\pm$ 0.075} & \makecell[r]{0.404 \\ $\pm$ 0.122} & \makecell[r]{0.474 \\ $\pm$ 0.102} & \makecell[r]{0.650 \\ $\pm$ 0.036} \\

    & ABMIL~\cite{ilse2018attention} & 0.9M & \makecell[r]{0.609 \\ $\pm$ 0.028} & \makecell[r]{0.656 \\ $\pm$ 0.055} & \makecell[r]{0.668 \\ $\pm$ 0.167} & \makecell[r]{0.606 \\ $\pm$ 0.044} & \makecell[r]{\textbf{0.712} \\ $\pm$ 0.057} & \makecell[r]{0.614 \\ $\pm$ 0.066} & \makecell[r]{0.595 \\ $\pm$ 0.073} & \makecell[r]{0.696 \\ $\pm$ 0.080} & \makecell[r]{0.650 \\ $\pm$ 0.098} & \makecell[r]{\underline{0.735} \\ $\pm$ 0.039} \\

    & TransMIL~\cite{shao2021transmil} & 2.7M & \makecell[r]{0.600 \\ $\pm$ 0.061} & \makecell[r]{0.663 \\ $\pm$ 0.058} & \makecell[r]{0.634 \\ $\pm$ 0.105} & \makecell[r]{0.587 \\ $\pm$ 0.041} & \makecell[r]{0.634 \\ $\pm$ 0.051} & \makecell[r]{0.587 \\ $\pm$ 0.076} & \makecell[r]{0.589 \\ $\pm$ 0.052} & \makecell[r]{0.636 \\ $\pm$ 0.121} & \makecell[r]{0.581 \\ $\pm$ 0.081} & \makecell[r]{0.704 \\ $\pm$ 0.052} \\

    \midrule

    \parbox[t]{2mm}{\multirow{2}{*}{\rotatebox[origin=c]{90}{{\textbf{Graph}}}}} 
    & Patch-GCN~\cite{chen2021whole} & 1.4M & \makecell[r]{0.597 \\ $\pm$ 0.022} & \makecell[r]{0.628 \\ $\pm$ 0.036} & \makecell[r]{0.634 \\ $\pm$ 0.121} & \makecell[r]{0.566 \\ $\pm$ 0.030} & \makecell[r]{0.648 \\ $\pm$ 0.074} & \makecell[r]{0.617 \\ $\pm$ 0.043} & \makecell[r]{0.590 \\ $\pm$ 0.063} & \makecell[r]{0.668 \\ $\pm$ 0.115} & \makecell[r]{0.563 \\ $\pm$ 0.048} & \makecell[r]{0.678 \\ $\pm$ 0.037} \\

    & WiKG~\cite{li2024dynamic} & 2.0M & \makecell[r]{0.638 \\ $\pm$ 0.030} & \makecell[r]{0.649 \\ $\pm$ 0.036} & \makecell[r]{0.722 \\ $\pm$ 0.069} & \makecell[r]{0.635 \\ $\pm$ 0.033} & \makecell[r]{0.657 \\ $\pm$ 0.067} & \makecell[r]{0.632 \\ $\pm$ 0.038} & \makecell[r]{\textbf{0.635} \\ $\pm$ 0.044} & \makecell[r]{0.661 \\ $\pm$ 0.112} & \makecell[r]{0.672 \\ $\pm$ 0.089} & \makecell[r]{0.723 \\ $\pm$ 0.036} \\

    \midrule

    \parbox[t]{2mm}{\multirow{5}{*}{\rotatebox[origin=c]{90}{{\textbf{Point Cloud}}}}} 
    & \makecell[c]{MeanPool (Cell)} & 1.5K & \makecell[r]{0.535 \\ $\pm$ 0.045} & \makecell[r]{0.573 \\ $\pm$ 0.070} & \makecell[r]{0.639 \\ $\pm$ 0.063} & \makecell[r]{0.584 \\ $\pm$ 0.027} & \makecell[r]{0.536 \\ $\pm$ 0.084} & \makecell[r]{0.552 \\ $\pm$ 0.068} & \makecell[r]{0.563 \\ $\pm$ 0.036} & \makecell[r]{0.647 \\ $\pm$ 0.045} & \makecell[r]{0.569 \\ $\pm$ 0.074} & \makecell[r]{0.594 \\ $\pm$ 0.034} \\

    & \makecell[c]{MaxPool (Cell)} & 1.5K & \makecell[r]{0.476 \\ $\pm$ 0.024} & \makecell[r]{0.571 \\ $\pm$ 0.055} & \makecell[r]{0.514 \\ $\pm$ 0.156} & \makecell[r]{0.548 \\ $\pm$ 0.054} & \makecell[r]{0.497 \\ $\pm$ 0.030} & \makecell[r]{0.535 \\ $\pm$ 0.035} & \makecell[r]{0.521 \\ $\pm$ 0.038} & \makecell[r]{0.544 \\ $\pm$ 0.065} & \makecell[r]{0.527 \\ $\pm$ 0.090} & \makecell[r]{0.563 \\ $\pm$ 0.095} \\

    & PointNet~\cite{qi2017pointnet} & 3.5M & \makecell[r]{0.633 \\ $\pm$ 0.025} & \makecell[r]{0.665 \\ $\pm$ 0.021} & \makecell[r]{0.732 \\ $\pm$ 0.044} & \makecell[r]{\underline{0.650} \\ $\pm$ 0.032} & \makecell[r]{0.649 \\ $\pm$ 0.035} & \makecell[r]{0.638 \\ $\pm$ 0.012} & \makecell[r]{0.612 \\ $\pm$ 0.021} & \makecell[r]{0.715 \\ $\pm$ 0.047} & \makecell[r]{0.682 \\ $\pm$ 0.075} & \makecell[r]{0.706 \\ $\pm$ 0.048} \\

    & PointNet++~\cite{qi2017pointnetpp} & 1.5M & \makecell[r]{0.613 \\ $\pm$ 0.038} & \makecell[r]{0.656 \\ $\pm$ 0.036} & \makecell[r]{0.743 \\ $\pm$ 0.028} & \makecell[r]{0.637 \\ $\pm$ 0.030} & \makecell[r]{0.627 \\ $\pm$ 0.045} & \makecell[r]{0.645 \\ $\pm$ 0.020} & \makecell[r]{\underline{0.634} \\ $\pm$ 0.036} & \makecell[r]{0.702 \\ $\pm$ 0.043} & \makecell[r]{0.633 \\ $\pm$ 0.055} & \makecell[r]{0.673 \\ $\pm$ 0.045} \\

    & PTv3~\cite{wu2024point} & 38.8M & \makecell[r]{0.553 \\ $\pm$ 0.039} & \makecell[r]{0.536 \\ $\pm$ 0.054} & \makecell[r]{0.616 \\ $\pm$ 0.069} & \makecell[r]{0.571 \\ $\pm$ 0.034} & \makecell[r]{0.498 \\ $\pm$ 0.037} & \makecell[r]{0.591 \\ $\pm$ 0.050} & \makecell[r]{0.579 \\ $\pm$ 0.041} & \makecell[r]{0.631 \\ $\pm$ 0.114} & \makecell[r]{0.560 \\ $\pm$ 0.023} & \makecell[r]{0.600 \\ $\pm$ 0.087} \\

    \midrule



    & \textbf{CCFormer (ours)} & 2.0M & \makecell[r]{\underline{0.645} \\ $\pm$ 0.031} & \makecell[r]{\underline{0.688} \\ $\pm$ 0.040} & \makecell[r]{\underline{0.753} \\ $\pm$ 0.069} & \makecell[r]{0.649 \\ $\pm$ 0.032} & \makecell[r]{0.658 \\ $\pm$ 0.052} & \makecell[r]{\underline{0.657} \\ $\pm$ 0.012} & \makecell[r]{0.633 \\ $\pm$ 0.019} & \makecell[r]{\underline{0.739} \\ $\pm$ 0.044} & \makecell[r]{\underline{0.687} \\ $\pm$ 0.062} & \makecell[r]{0.693 \\ $\pm$ 0.049}\\

    & \makecell[c]{\textbf{CCFormer (ours)} \\ \textbf{+ MeanPool (Patch)}} & 2.7M & \makecell[r]{\textbf{0.664} \\ $\pm$ 0.019} & \makecell[r]{\textbf{0.704} \\ $\pm$ 0.065} & \makecell[r]{\textbf{0.756} \\ $\pm$ 0.076} & \makecell[r]{\textbf{0.652} \\ $\pm$ 0.032} & \makecell[r]{\underline{0.696} \\ $\pm$ 0.038} & \makecell[r]{\textbf{0.660} \\ $\pm$ 0.018} & \makecell[r]{0.626 \\ $\pm$ 0.017} & \makecell[r]{\textbf{0.741} \\ $\pm$ 0.066} & \makecell[r]{\textbf{0.704} \\ $\pm$ 0.062} & \makecell[r]{\textbf{0.738} \\ $\pm$ 0.052}\\

    \bottomrule
    
    \end{tabular}
    
    }
    \caption{Comparison of survival prediction with SOTA methods in C-Index ($\uparrow$).
    }
    \label{tb:survival_prediction}
    
\end{table*}

The semantic-spatial aware filter comprehensively considers semantic similarity and spatial distance of cells within the same group.
For each group, the coordinates of the group anchor and the mean feature of the cells within the group are as references. Then, a similarity score is computed for each cell:
\begin{equation}
\small
\label{eq:compute_mask}
\begin{split}
    S_{sim}^{(i)} = \exp{(-\Vert c^{(i)} - c_{ref} \Vert)}\frac{(f^{(i)}_{cell})^Tf_{ref}}{N_{dim}},
\end{split}
\end{equation}
where $S_{sim}^{(i)}$ is the similarity score of the $i$-th cell, $f_{ref}$ denotes the the mean feature, $c_{ref}$ denotes the coordinate of the group kernel, and $\Vert \cdot \Vert$ denotes the euclidean distance.
We introduce a threshold $\lambda_{sim}$ to generate filter $M$.
Specifically, if $S_{sim}^{(i)} < \lambda_{sim}$, $M^{(i)} = 0$ and the $i$-th cell is marked for discard.

For each cell, we calculate attention weights with respect to other cells within the same group to update its feature.
The attention is implemented as vector attention~\cite{zhao2021point}.
Moreover, the positional relationships among cells are critical spatial information. Therefore, we incorporate relative coordinates into our calculations.
Assuming that the $i$-th cell within the group $G$, the information interaction between this cell and other cells is defined as follows:
\begin{equation}
\small
\label{eq:att}
\begin{split}
    S_{att}^{(i, j)} &= M^{(j)}\mathcal{F}_{att}(\mathcal{W}_{q}f_{cell}^{(i)} - \mathcal{W}_{k}f_{cell}^{(j)} + d^{(i, j)}), \\
    (f_{cell}^{(i)})^\prime &= \sum_{G} \delta(S_{att}^{(i, j)})(\mathcal{W}_{v}f_{cell}^{(j)} + d^{(i, j)}), \\
    d^{(i, j)} &= c^{(i)} - c^{(j)},
\end{split}
\end{equation}
where $S_{att}^{(i, j)}$, and $d^{(i, j)}$ are the attention vector and distance between the $i$-th cell and $j$-th cell, respectively, $\mathcal{F}_{att}$ is a Multilayer Perceptron (MLP), $(f_{cell}^{(i)})^\prime$ is the updated feature of the $i$-th cell, $\delta$ denote the softmax and normalization for attention vectors, and $\mathcal{W}_{\cdot}$ denotes a linear projection layer.
Cells can perceive the spatial distribution by stacking layers as described in Equation \ref{eq:att}.



HSP further introduces a hierarchical architecture to model the cell spatial distribution in a bottom-up manner, as detailed in the pseudocode found in the appendix.
HSP consists of multiple levels, each of which models the local spatial spatial distribution of cells at a specific scale.
Higher-level features are derived from lower-level features by mean aggregation and are subsequently re-grouped and undergo attention to model cell spatial distribution over a larger region.
Finally, the feature of the WSI is the maximum aggregation of features at the last level.

%% file: sec/6_exps.tex
\section{Experiments}\label{sec:exps}

\subsection{Experiment Settings}

\noindent \textbf{Datasets.}
We conduct extensive experiments across all cancer types included in the WSI-Cell5B.
Specifically, the experiments encompass two critical tasks in WSI analysis, including survival prediction and cancer staging.

\begin{figure*}[t]
    \centering
    \includegraphics[width=0.99\linewidth]{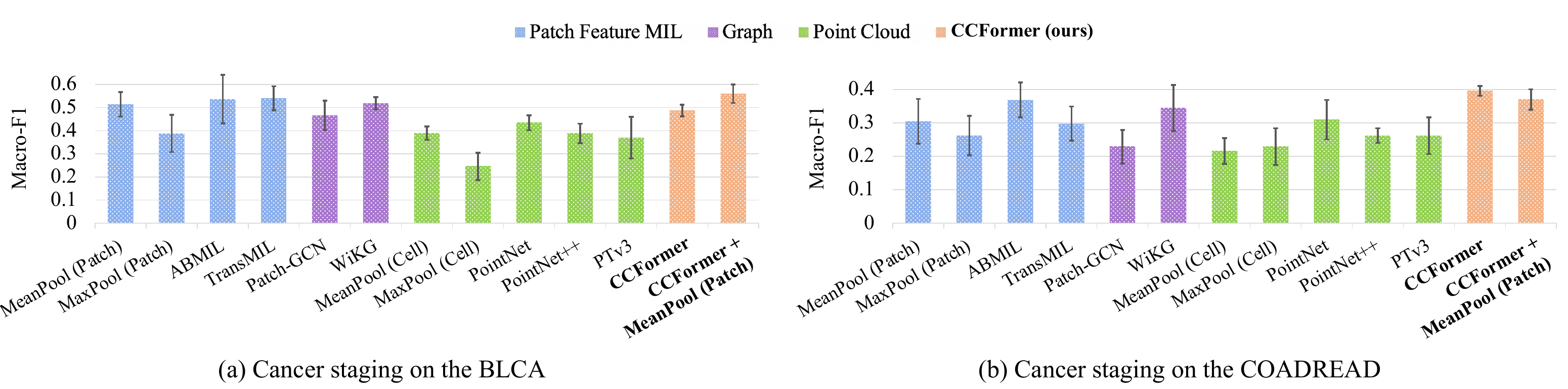}
    \caption{
    Comparison of cancer staging with SOTA methods on (a) BLCA and (b) COADREAD in Macro-F1 ($\uparrow$).
    }
    \label{fig:staging}
\end{figure*} 

\noindent \textbf{Metrics.}
In our experiments, we follow the usual practice to evaluate the performance of survival prediction~\cite{chen2021whole, nakhli2023sparse} and cancer staging~\cite{li2024dynamic, chan2023histopathology} by C-Index and Macro-F1 with 5-fold cross-validation, respectively.

\noindent \textbf{Baselines.}
Global mean pooling (MeanPool) and global max pooling (MaxPool) are employed as baselines.
In addition, we compare CCFormer with SOTA WSI analysis methods.
For MIL-based methods, we compare with ABMIL~\cite{ilse2018attention} and TransMIL~\cite{shao2021transmil}.
As MIL methods are highly influenced by the pre-trained feature extractor, we utilize the pre-trained UNI~\cite{chen2024towards}, a SOTA self-supervised model for pathology, to extract patch features.
For graph-based methods, we compare with Patch-GCN~\cite{chen2021whole} and WiKG~\cite{li2024dynamic}.
In addition, models for learning point clouds can be adaptively applied to cell clouds. Thus, we compare CCFormer with SOTA point cloud methods, including PointNet~\cite{qi2017pointnet}, PointNet++~\cite{qi2017pointnetpp}, and Point Transformer v3 (PTv3)~\cite{wu2024point}.






\noindent \textbf{Implementation Details.}
 Please refer to the supplement.

 \begin{table}[t]
    \centering
    \small
    \renewcommand{\tabcolsep}{24pt}
    \renewcommand{\arraystretch}{1}
    \resizebox{\linewidth}{!}{
    \begin{tabular}{ccc|c}
    \toprule
    
    Type & $F_{ld}$ & $F_{gd}$ & C-Index \\
    \midrule
    \ding{51} & \ding{55} & \ding{55} & 0.678 $\pm$ 0.062 \\
    \ding{51} & \ding{51} & \ding{55} & 0.722 $\pm$ 0.075 \\
    \ding{51} & \ding{55} & \ding{51} & 0.712 $\pm$ 0.042 \\
    \ding{55} & \ding{51} & \ding{51} & 0.729 $\pm$ 0.085 \\
    \ding{51} & \ding{51} & \ding{51} & \textbf{0.739} $\pm$ 0.044 \\
    \bottomrule
    \end{tabular}
    }

    \caption{Ablation study of Neighboring Information Embedding (NIE) on PAAD in C-Index ($\uparrow$). 
    }
    \label{tb:ablation_feature}
\end{table}

\subsection{Main Results}

Table \ref{tb:survival_prediction} and Figure \ref{fig:staging} report the results of survival prediction and cancer staging, respectively.
Point cloud methods and CCFormer achieve competitive results with MIL-based and graph-based methods.
Results indicate that patient survival risk and cancer stages are closely related to cell spatial distribution, which aids in the analysis of WSIs and further enhances the accuracy of downstream tasks such as survival prediction and cancer staging.

\begin{figure}[t]
    \centering
    \includegraphics[width=0.90\linewidth]{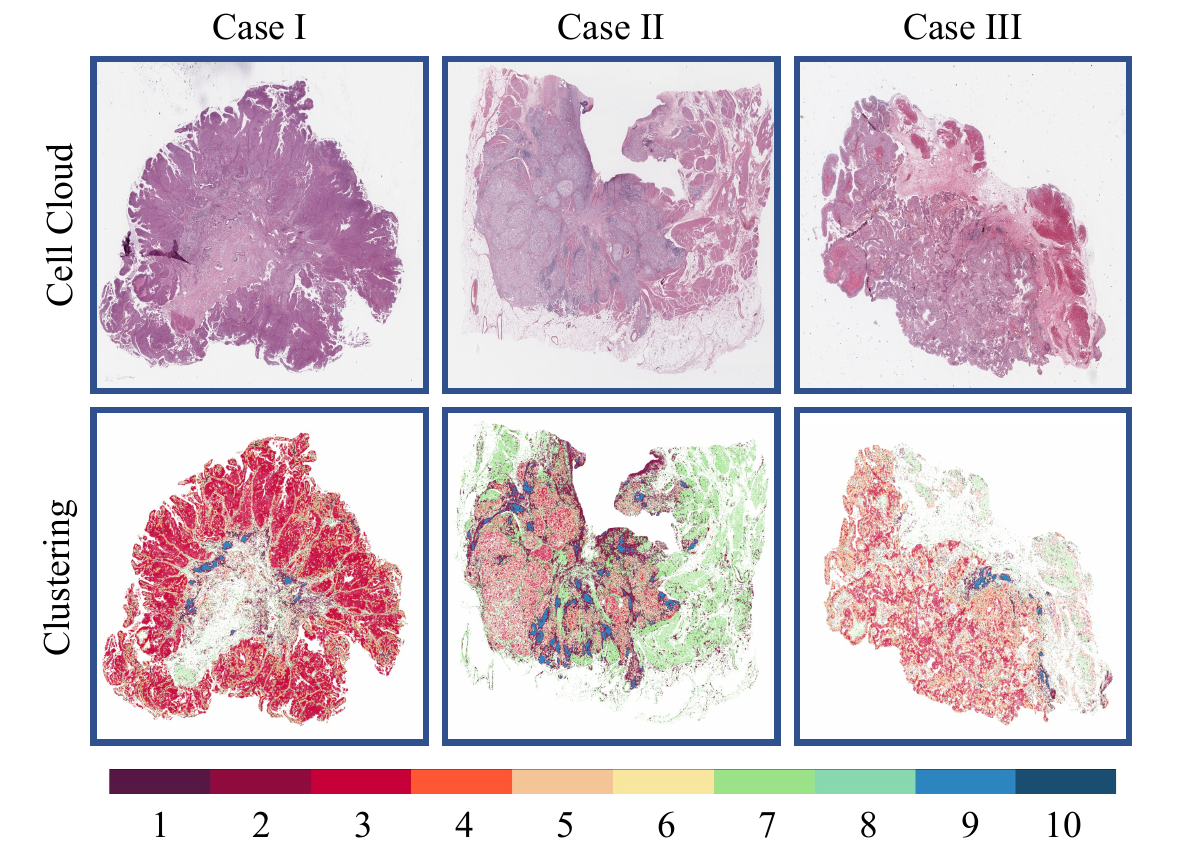}
    \caption{
    {Visualization of clustering based on $F_{ld}$ and $F_{gd}$.}
    }
    \label{fig:vis_feature}
\end{figure} 

\noindent \textbf{Survival Prediction.}
As shown in Table \ref{tb:survival_prediction}, CCFormer outperforms other SOTA methods on most cancer types.
CCFormer achieves the highest C-Index with improvements of 1\% to 4\% on the BLCA, BRCA, COADREAD, LUAD, PAAD, and STAD.
Besides, CCFormer achieves a C-Index similar to that of the best methods on the HNSC and LUSC, with differences less than or equal to 0.002.

However, modeling cell spatial distribution is insufficient for survival prediction in some cancer types.
As evidence, CCFormer and point cloud methods perform inferior to patch-based methods on the KIRC and UCEC.
Therefore, we further combine CCFormer and global mean pooling of patches to validate the performance of combining cell spatial distribution with WSI image features in survival prediction.
As shown in Table \ref{tb:survival_prediction}, the combination method achieves a better C-Index on all datasets than the global mean pooling.
Compared to CCFormer, the combined method achieves better C-Index on all datasets except LUSC, in which the global mean pooling exhibits significantly poor performance, leading to a decrease in C-Index.

\noindent \textbf{Cancer Staging.}
Figure \ref{fig:staging} (a) reports the results of cancer staging on the BLCA.
CCFormer outperforms PointNet, PointNet++, and PTv3 with 12\%, 26\%, and 32\%, respectively.
However, merely learning the cell spatial distribution can not completely distinguish the stages of cancer.
Specifically, the methods based on patch features exhibit a significantly higher Macro-F1.
By integrating features of cell spatial distribution and WSI appearance, the combination of CCFormer and the global mean pooling achieves higher Micro-F1 and outperforms the best patch-based method with 4\%.

Figure \ref{fig:staging} (b) reports the results of cancer staging on the COADREAD.
CCFormer outperforms PointNet, PointNet++, and PTv3 with 28\%, 51\%, and 51\%, respectively.
Compared to SOTA patch-based methods, including ABMIL, TransMIL, Patch-GCN, and WiKG, CCFormer achieves significant improvement with 7\%, 33\%, 73\%, and 15\%, respectively.
The experiments show that the global mean pooling can not provide effective WSI features for cancer staging on the COADREAD, leading to a decrease in Macro-F1 after combining CCFormer with MeanPool.

\subsection{Ablation Studies}

\noindent \textbf{Neighboring Information Embedding.}
Table \ref{tb:ablation_feature} reports the results of ablation studies on NIE.
Both $F_{ld}$ and $F_{gd}$ individually improve the C-Index of survival prediction on PAAD.
The combined use of $F_{ld}$ and $F_{gd}$ further enhances the accuracy.
Specifically, method with $F_{ld}$, $F_{gd}$, and $F_{ld}+F_{gd}$ outperform the baseline with 6\%, 5\%, and 9\%, respectively.
By default, the one-hot embedding of cell types is utilized as input in CCFormer.
Ablation studies also show the importance of modeling neighboring cell spatial distributions that CCFormer achieves a better C-Index than the baseline even without cell types as input.

We cluster cells based on $F_{ld}$ and $F_{gd}$ via K-Means to visualize NIE.
As shown in Figure \ref{fig:vis_feature}, cells with different neighboring cell spatial distributions are distinguished.
Specifically, cancer cell clusters, immune cell clusters, mixed cell regions, and outlier cells are identified.

\begin{figure}[t]
    \centering
    \includegraphics[width=0.90\linewidth]{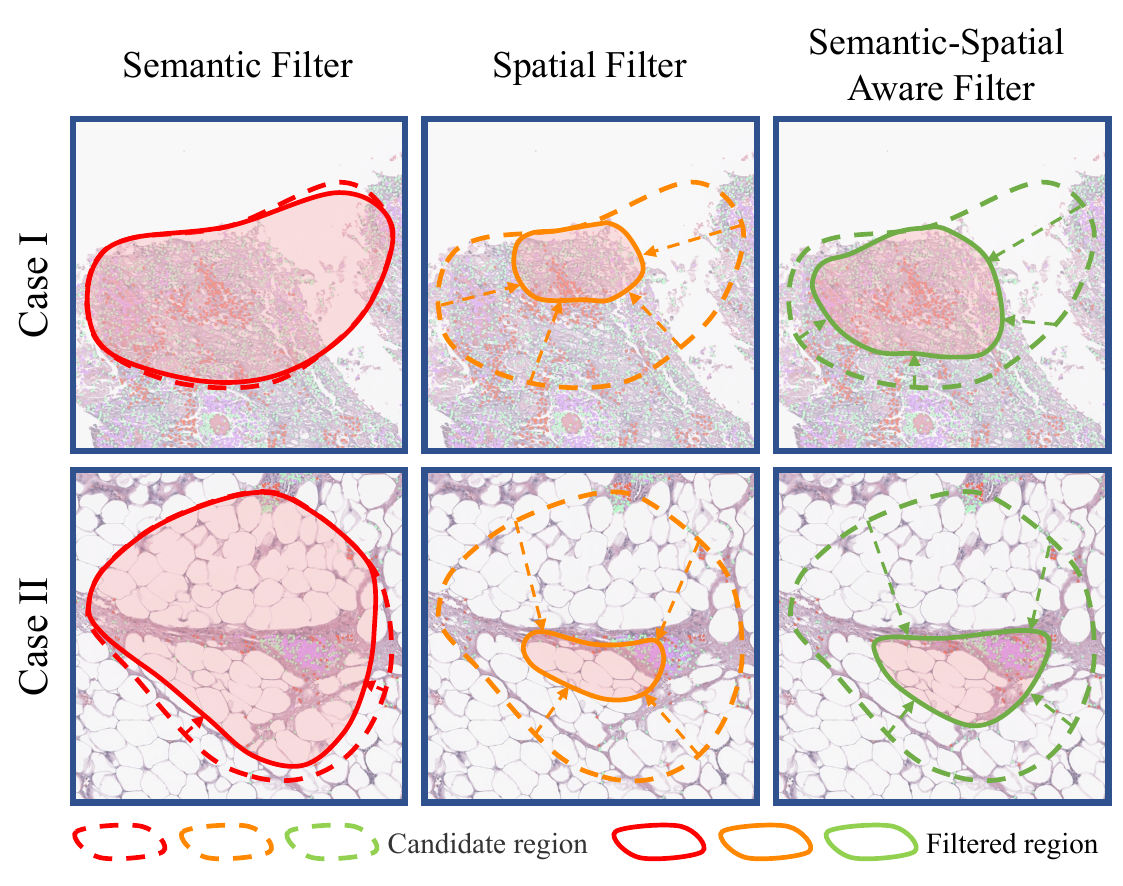}
    \caption{Visualization of the semantic-spatial aware filter.
    }
    \label{fig:vis_filter}
\end{figure} 

\noindent \textbf{Hierarchical Spatial Perception.}
Table \ref{tb:ablation_hsp} reports the results of ablation studies on HSP.
Compared to the baseline that generates groups only based on spatial information, introducing cell types as additional semantic information into the grouping can generate more reasonable groups and improve the C-Index of survival prediction.

Ablation studies also show that filtering cells within each group based on spatial and semantics improves performance.
Specifically, HSP with spatial filtering and HSP that filters cells using both semantic and spatial information outperform the baseline with 1\% and 2\%, respectively.
As semantic filtering tends to exclude heterogeneous cells, HSP with a semantic filter forcefully filters out cells that significantly differ from the mean feature, leading to incorrect modeling of cell spatial distribution.
In addition, the filter with a large threshold removes lots of important cells, resulting in a decrease in C-Index.

Figure \ref{fig:vis_filter} shows the visualization of three types of filters.
The semantic filter incorrectly filters important heterogeneous cells. 
In addition, due to the lack of semantic awareness, the spatial filter fails to fully preserve semantic relevance cells.
The semantic-spatial aware filter retains the primary cell clusters while striving to remove irrelevant cells as much as possible.



\begin{table}[t]
    \centering
    \small
    \renewcommand{\tabcolsep}{5pt}
    \renewcommand{\arraystretch}{1}
    \resizebox{\linewidth}{!}{
    \begin{tabular}{c|cc|c}
    \toprule

    Grouping & Filter Type & $\lambda_{sim}$ & C-Index \\
    \midrule
    Spatial & None & - & 0.727 $\pm$ 0.068 \\
    Spatial + Semantic & None & - & 0.728 $\pm$ 0.069 \\
    Spatial + Semantic & Semantic & 0.5 & 0.720 $\pm$ 0.064 \\
    Spatial + Semantic & Spatial & 0.5 & 0.736 $\pm$ 0.078 \\
    Spatial + Semantic & Spatial + Semantic & 0.8 & 0.709 $\pm$ 0.095 \\
    Spatial + Semantic & Spatial + Semantic & 0.5 & \textbf{0.739} $\pm$ 0.044\\
    \bottomrule
    \end{tabular}
    }
    \caption{Ablation study of Hierarchical Spatial Perception (HSP) on PAAD in C-Idex ($\uparrow$).
    }
    \label{tb:ablation_hsp}
\end{table}

%% file: sec/7_conclusion.tex
\section{Conclusion}
\label{sec:conclusion}

In this paper, we propose WSI-Cell5B, the first large-scale WSI dataset to integrate cell-level annotations with WSIs, comprising 6,998 WSIs of 11 cancers and more than 5 billion cell-level annotations.
To reduce the cost of cell-level annotations, we further propose a weakly supervised label refinement method based on foundation models.
We argue the collection of cells within a WSI can be regarded as a cell cloud and propose a novel hierarchical Cell Cloud Transformer (CCFormer) to model the spatial distribution of cells.
We propose a novel Neighboring Information Embedding (NIE) to embed the neighborhood cell distribution at the cell level and a novel Hierarchical Spatial Perception (HSP) to progressively perceive and aggregate cell spatial distribution information.
The clinical analysis validates that WSI-Cell5B can be directly used to construct effective clinical indicators. 
In addition, extensive experiments verify that cell cloud is an effective representation of slide and CCFormer outperforms other SOTA methods.
This work provides a new insight for WSI analysis from the cell cloud perspective, marking a significant milestone in the advancement of computational pathology.

%% file: main.bbl
\begin{thebibliography}{60}
\providecommand{\natexlab}[1]{#1}
\providecommand{\url}[1]{\texttt{#1}}
\expandafter\ifx\csname urlstyle\endcsname\relax
  \providecommand{\doi}[1]{doi: #1}\else
  \providecommand{\doi}{doi: \begingroup \urlstyle{rm}\Url}\fi

\bibitem[Bandi et~al.(2018)Bandi, Geessink, Manson, Van~Dijk, Balkenhol, Hermsen, Bejnordi, Lee, Paeng, Zhong, et~al.]{bandi2018detection}
Peter Bandi, Oscar Geessink, Quirine Manson, Marcory Van~Dijk, Maschenka Balkenhol, Meyke Hermsen, Babak~Ehteshami Bejnordi, Byungjae Lee, Kyunghyun Paeng, Aoxiao Zhong, et~al.
\newblock From detection of individual metastases to classification of lymph node status at the patient level: the camelyon17 challenge.
\newblock \emph{IEEE transactions on medical imaging}, 38\penalty0 (2):\penalty0 550--560, 2018.

\bibitem[Bulten et~al.(2022)Bulten, Kartasalo, Chen, Str{\"o}m, Pinckaers, Nagpal, Cai, Steiner, Van~Boven, Vink, et~al.]{bulten2022artificial}
Wouter Bulten, Kimmo Kartasalo, Po-Hsuan~Cameron Chen, Peter Str{\"o}m, Hans Pinckaers, Kunal Nagpal, Yuannan Cai, David~F Steiner, Hester Van~Boven, Robert Vink, et~al.
\newblock Artificial intelligence for diagnosis and gleason grading of prostate cancer: the panda challenge.
\newblock \emph{Nature medicine}, 28\penalty0 (1):\penalty0 154--163, 2022.

\bibitem[Chan et~al.(2023)Chan, Cendra, Ma, Yin, and Yu]{chan2023histopathology}
Tsai~Hor Chan, Fernando~Julio Cendra, Lan Ma, Guosheng Yin, and Lequan Yu.
\newblock Histopathology whole slide image analysis with heterogeneous graph representation learning.
\newblock In \emph{Proceedings of the IEEE/CVF Conference on Computer Vision and Pattern Recognition}, pages 15661--15670, 2023.

\bibitem[Chen et~al.(2021)Chen, Lu, Shaban, Chen, Chen, Williamson, and Mahmood]{chen2021whole}
Richard~J Chen, Ming~Y Lu, Muhammad Shaban, Chengkuan Chen, Tiffany~Y Chen, Drew~FK Williamson, and Faisal Mahmood.
\newblock Whole slide images are 2d point clouds: Context-aware survival prediction using patch-based graph convolutional networks.
\newblock In \emph{Medical Image Computing and Computer Assisted Intervention {\textendash} {MICCAI} 2021}, pages 339--349. Springer International Publishing, 2021.

\bibitem[Chen et~al.(2022)Chen, Chen, Li, Chen, Trister, Krishnan, and Mahmood]{chen2022scaling}
Richard~J Chen, Chengkuan Chen, Yicong Li, Tiffany~Y Chen, Andrew~D Trister, Rahul~G Krishnan, and Faisal Mahmood.
\newblock Scaling vision transformers to gigapixel images via hierarchical self-supervised learning.
\newblock In \emph{Proceedings of the IEEE/CVF Conference on Computer Vision and Pattern Recognition}, pages 16144--16155, 2022.

\bibitem[Chen et~al.(2024)Chen, Ding, Lu, Williamson, Jaume, Song, Chen, Zhang, Shao, Shaban, et~al.]{chen2024towards}
Richard~J Chen, Tong Ding, Ming~Y Lu, Drew~FK Williamson, Guillaume Jaume, Andrew~H Song, Bowen Chen, Andrew Zhang, Daniel Shao, Muhammad Shaban, et~al.
\newblock Towards a general-purpose foundation model for computational pathology.
\newblock \emph{Nature Medicine}, 30\penalty0 (3):\penalty0 850--862, 2024.

\bibitem[Chen et~al.(2023)Chen, Liu, Zhang, Qi, and Jia]{chen2023largekernel3d}
Yukang Chen, Jianhui Liu, Xiangyu Zhang, Xiaojuan Qi, and Jiaya Jia.
\newblock Largekernel3d: Scaling up kernels in 3d sparse cnns.
\newblock In \emph{Proceedings of the IEEE/CVF Conference on Computer Vision and Pattern Recognition}, pages 13488--13498, 2023.

\bibitem[Choy et~al.(2019)Choy, Gwak, and Savarese]{choy20194d}
Christopher Choy, JunYoung Gwak, and Silvio Savarese.
\newblock 4d spatio-temporal convnets: Minkowski convolutional neural networks.
\newblock In \emph{Proceedings of the IEEE/CVF conference on computer vision and pattern recognition}, pages 3075--3084, 2019.

\bibitem[Corredor et~al.(2019)Corredor, Wang, Zhou, Lu, Fu, Syrigos, Rimm, Yang, Romero, Schalper, et~al.]{corredor2019spatial}
Germ{\'a}n Corredor, Xiangxue Wang, Yu Zhou, Cheng Lu, Pingfu Fu, Konstantinos Syrigos, David~L Rimm, Michael Yang, Eduardo Romero, Kurt~A Schalper, et~al.
\newblock Spatial architecture and arrangement of tumor-infiltrating lymphocytes for predicting likelihood of recurrence in early-stage non--small cell lung cancer.
\newblock \emph{Clinical cancer research}, 25\penalty0 (5):\penalty0 1526--1534, 2019.

\bibitem[Dai et~al.(2017)Dai, Chang, Savva, Halber, Funkhouser, and Nie{\ss}ner]{dai2017scannet}
Angela Dai, Angel~X. Chang, Manolis Savva, Maciej Halber, Thomas Funkhouser, and Matthias Nie{\ss}ner.
\newblock Scannet: Richly-annotated 3d reconstructions of indoor scenes.
\newblock In \emph{Proc. Computer Vision and Pattern Recognition (CVPR), IEEE}, 2017.

\bibitem[Dyrskj{\o}t et~al.(2023)Dyrskj{\o}t, Hansel, Efstathiou, Knowles, Galsky, Teoh, and Theodorescu]{dyrskjot2023bladder}
Lars Dyrskj{\o}t, Donna~E Hansel, Jason~A Efstathiou, Margaret~A Knowles, Matthew~D Galsky, Jeremy Teoh, and Dan Theodorescu.
\newblock Bladder cancer.
\newblock \emph{Nature Reviews Disease Primers}, 9\penalty0 (1):\penalty0 58, 2023.

\bibitem[Frank et~al.(2002)Frank, Blute, Cheville, Lohse, Weaver, and Zincke]{frank2002outcome}
IGOR Frank, Michael~L Blute, John~C Cheville, Christine~M Lohse, Amy~L Weaver, and Horst Zincke.
\newblock An outcome prediction model for patients with clear cell renal cell carcinoma treated with radical nephrectomy based on tumor stage, size, grade and necrosis: the ssign score.
\newblock \emph{The Journal of urology}, 168\penalty0 (6):\penalty0 2395--2400, 2002.

\bibitem[Gamper et~al.(2020)Gamper, Koohbanani, Benes, Graham, Jahanifar, Khurram, Azam, Hewitt, and Rajpoot]{gamper2020pannuke}
Jevgenij Gamper, Navid~Alemi Koohbanani, Ksenija Benes, Simon Graham, Mostafa Jahanifar, Syed~Ali Khurram, Ayesha Azam, Katherine Hewitt, and Nasir Rajpoot.
\newblock Pannuke dataset extension, insights and baselines.
\newblock \emph{arXiv preprint arXiv:2003.10778}, 2020.

\bibitem[Graham et~al.(2018)Graham, Engelcke, and Van Der~Maaten]{graham20183d}
Benjamin Graham, Martin Engelcke, and Laurens Van Der~Maaten.
\newblock 3d semantic segmentation with submanifold sparse convolutional networks.
\newblock In \emph{Proceedings of the IEEE conference on computer vision and pattern recognition}, pages 9224--9232, 2018.

\bibitem[Graham et~al.(2019)Graham, Vu, Raza, Azam, Tsang, Kwak, and Rajpoot]{graham2019hover}
Simon Graham, Quoc~Dang Vu, Shan E~Ahmed Raza, Ayesha Azam, Yee~Wah Tsang, Jin~Tae Kwak, and Nasir Rajpoot.
\newblock Hover-net: Simultaneous segmentation and classification of nuclei in multi-tissue histology images.
\newblock \emph{Medical image analysis}, 58:\penalty0 101563, 2019.

\bibitem[Graham et~al.(2021)Graham, Jahanifar, Azam, Nimir, Tsang, Dodd, Hero, Sahota, Tank, Benes, et~al.]{graham2021lizard}
Simon Graham, Mostafa Jahanifar, Ayesha Azam, Mohammed Nimir, Yee-Wah Tsang, Katherine Dodd, Emily Hero, Harvir Sahota, Atisha Tank, Ksenija Benes, et~al.
\newblock Lizard: A large-scale dataset for colonic nuclear instance segmentation and classification.
\newblock In \emph{Proceedings of the IEEE/CVF international conference on computer vision}, pages 684--693, 2021.

\bibitem[Grossman et~al.(2016)Grossman, Heath, Ferretti, Varmus, Lowy, Kibbe, and Staudt]{grossman2016toward}
Robert~L Grossman, Allison~P Heath, Vincent Ferretti, Harold~E Varmus, Douglas~R Lowy, Warren~A Kibbe, and Louis~M Staudt.
\newblock Toward a shared vision for cancer genomic data.
\newblock \emph{New England Journal of Medicine}, 375\penalty0 (12):\penalty0 1109--1112, 2016.

\bibitem[He et~al.(2016)He, Zhang, Ren, and Sun]{he2016deep}
Kaiming He, Xiangyu Zhang, Shaoqing Ren, and Jian Sun.
\newblock Deep residual learning for image recognition.
\newblock In \emph{Proceedings of the IEEE conference on computer vision and pattern recognition}, pages 770--778, 2016.

\bibitem[Ikezogwo et~al.(2024)Ikezogwo, Seyfioglu, Ghezloo, Geva, Sheikh~Mohammed, Anand, Krishna, and Shapiro]{ikezogwo2024quilt}
Wisdom Ikezogwo, Saygin Seyfioglu, Fatemeh Ghezloo, Dylan Geva, Fatwir Sheikh~Mohammed, Pavan~Kumar Anand, Ranjay Krishna, and Linda Shapiro.
\newblock Quilt-1m: One million image-text pairs for histopathology.
\newblock \emph{Advances in neural information processing systems}, 36, 2024.

\bibitem[Ilse et~al.(2018)Ilse, Tomczak, and Welling]{ilse2018attention}
Maximilian Ilse, Jakub Tomczak, and Max Welling.
\newblock Attention-based deep multiple instance learning.
\newblock In \emph{International conference on machine learning}, pages 2127--2136. PMLR, 2018.

\bibitem[Jaume et~al.(2024)Jaume, Vaidya, Chen, Williamson, Liang, and Mahmood]{jaume2023modeling}
Guillaume Jaume, Anurag Vaidya, Richard Chen, Drew Williamson, Paul Liang, and Faisal Mahmood.
\newblock Modeling dense multimodal interactions between biological pathways and histology for survival prediction.
\newblock \emph{Proceedings of the IEEE/CVF Conference on Computer Vision and Pattern Recognition (CVPR)}, 2024.

\bibitem[Kingma(2014)]{kingma2014adam}
Diederik~P Kingma.
\newblock Adam: A method for stochastic optimization.
\newblock \emph{arXiv preprint arXiv:1412.6980}, 2014.

\bibitem[Kumar et~al.(2017)Kumar, Verma, Sharma, Bhargava, Vahadane, and Sethi]{kumar2017dataset}
Neeraj Kumar, Ruchika Verma, Sanuj Sharma, Surabhi Bhargava, Abhishek Vahadane, and Amit Sethi.
\newblock A dataset and a technique for generalized nuclear segmentation for computational pathology.
\newblock \emph{IEEE transactions on medical imaging}, 36\penalty0 (7):\penalty0 1550--1560, 2017.

\bibitem[Lang et~al.(2019)Lang, Vora, Caesar, Zhou, Yang, and Beijbom]{lang2019pointpillars}
Alex~H Lang, Sourabh Vora, Holger Caesar, Lubing Zhou, Jiong Yang, and Oscar Beijbom.
\newblock Pointpillars: Fast encoders for object detection from point clouds.
\newblock In \emph{Proceedings of the IEEE/CVF conference on computer vision and pattern recognition}, pages 12697--12705, 2019.

\bibitem[Li et~al.(2024)Li, Chen, Chu, Sun, Guan, Han, and He]{li2024dynamic}
Jiawen Li, Yuxuan Chen, Hongbo Chu, Qiehe Sun, Tian Guan, Anjia Han, and Yonghong He.
\newblock Dynamic graph representation with knowledge-aware attention for histopathology whole slide image analysis.
\newblock In \emph{Proceedings of the IEEE/CVF Conference on Computer Vision and Pattern Recognition}, pages 11323--11332, 2024.

\bibitem[Liang et~al.(2024)Liang, Zhou, Xu, Zhu, Zou, Ye, Tan, and Bai]{liang2024pointmamba}
Dingkang Liang, Xin Zhou, Wei Xu, Xingkui Zhu, Zhikang Zou, Xiaoqing Ye, Xiao Tan, and Xiang Bai.
\newblock Pointmamba: A simple state space model for point cloud analysis.
\newblock \emph{arXiv preprint arXiv:2402.10739}, 2024.

\bibitem[Lin et~al.(2023)Lin, Yu, Hu, Xu, and Chen]{lin2023interventional}
Tiancheng Lin, Zhimiao Yu, Hongyu Hu, Yi Xu, and Chang-Wen Chen.
\newblock Interventional bag multi-instance learning on whole-slide pathological images.
\newblock In \emph{Proceedings of the IEEE/CVF Conference on Computer Vision and Pattern Recognition}, pages 19830--19839, 2023.

\bibitem[Liu et~al.(2018)Liu, Lichtenberg, Hoadley, Poisson, Lazar, Cherniack, Kovatich, Benz, Levine, Lee, et~al.]{liu2018integrated}
Jianfang Liu, Tara Lichtenberg, Katherine~A Hoadley, Laila~M Poisson, Alexander~J Lazar, Andrew~D Cherniack, Albert~J Kovatich, Christopher~C Benz, Douglas~A Levine, Adrian~V Lee, et~al.
\newblock An integrated tcga pan-cancer clinical data resource to drive high-quality survival outcome analytics.
\newblock \emph{Cell}, 173\penalty0 (2):\penalty0 400--416, 2018.

\bibitem[Liu et~al.(2019)Liu, Jiang, He, Chen, Liu, Gao, and Han]{liu2019variance}
Liyuan Liu, Haoming Jiang, Pengcheng He, Weizhu Chen, Xiaodong Liu, Jianfeng Gao, and Jiawei Han.
\newblock On the variance of the adaptive learning rate and beyond.
\newblock \emph{arXiv preprint arXiv:1908.03265}, 2019.

\bibitem[Lu et~al.(2021)Lu, Williamson, Chen, Chen, Barbieri, and Mahmood]{lu2021data}
Ming~Y Lu, Drew~FK Williamson, Tiffany~Y Chen, Richard~J Chen, Matteo Barbieri, and Faisal Mahmood.
\newblock Data-efficient and weakly supervised computational pathology on whole-slide images.
\newblock \emph{Nature Biomedical Engineering}, 5\penalty0 (6):\penalty0 555--570, 2021.

\bibitem[Lu et~al.(2023)Lu, Chen, Williamson, Chen, Ikamura, Gerber, Liang, Le, Ding, Parwani, et~al.]{pathchat}
Ming~Y Lu, Bowen Chen, Drew~FK Williamson, Richard~J Chen, Kenji Ikamura, Georg Gerber, Ivy Liang, Long~Phi Le, Tong Ding, Anil~V Parwani, et~al.
\newblock A foundational multimodal vision language ai assistant for human pathology.
\newblock \emph{arXiv preprint arXiv:2312.07814}, 2023.

\bibitem[Lu et~al.(2024{\natexlab{a}})Lu, Chen, Williamson, Chen, Liang, Ding, Jaume, Odintsov, Le, Gerber, et~al.]{lu2024avisionlanguage}
Ming~Y Lu, Bowen Chen, Drew~FK Williamson, Richard~J Chen, Ivy Liang, Tong Ding, Guillaume Jaume, Igor Odintsov, Long~Phi Le, Georg Gerber, et~al.
\newblock A visual-language foundation model for computational pathology.
\newblock \emph{Nature Medicine}, 30:\penalty0 863–874, 2024{\natexlab{a}}.

\bibitem[Lu et~al.(2024{\natexlab{b}})Lu, Chen, Williamson, Chen, Liang, Ding, Jaume, Odintsov, Le, Gerber, et~al.]{lu2024visual}
Ming~Y Lu, Bowen Chen, Drew~FK Williamson, Richard~J Chen, Ivy Liang, Tong Ding, Guillaume Jaume, Igor Odintsov, Long~Phi Le, Georg Gerber, et~al.
\newblock A visual-language foundation model for computational pathology.
\newblock \emph{Nature Medicine}, 30\penalty0 (3):\penalty0 863--874, 2024{\natexlab{b}}.

\bibitem[Ma et~al.(2022)Ma, Qin, You, Ran, and Fu]{ma2022rethinking}
Xu Ma, Can Qin, Haoxuan You, Haoxi Ran, and Yun Fu.
\newblock Rethinking network design and local geometry in point cloud: A simple residual mlp framework.
\newblock \emph{ICLR}, 2022.

\bibitem[Maturana and Scherer(2015)]{maturana2015voxnet}
Daniel Maturana and Sebastian Scherer.
\newblock Voxnet: A 3d convolutional neural network for real-time object recognition.
\newblock In \emph{2015 IEEE/RSJ international conference on intelligent robots and systems (IROS)}, pages 922--928. IEEE, 2015.

\bibitem[Nakhli et~al.(2023)Nakhli, Moghadam, Mi, Farahani, Baras, Gilks, and Bashashati]{nakhli2023sparse}
Ramin Nakhli, Puria~Azadi Moghadam, Haoyang Mi, Hossein Farahani, Alexander Baras, Blake Gilks, and Ali Bashashati.
\newblock Sparse multi-modal graph transformer with shared-context processing for representation learning of giga-pixel images.
\newblock In \emph{Proceedings of the IEEE/CVF Conference on Computer Vision and Pattern Recognition}, pages 11547--11557, 2023.

\bibitem[Page et~al.(2023)Page, Broeckx, Jahangir, Verbandt, Gupta, Thagaard, Khiroya, Kos, Abduljabbar, Acosta~Haab, et~al.]{page2023spatial}
David~B Page, Glenn Broeckx, Chowdhury~Arif Jahangir, Sara Verbandt, Rajarsi~R Gupta, Jeppe Thagaard, Reena Khiroya, Zuzana Kos, Khalid Abduljabbar, Gabriela Acosta~Haab, et~al.
\newblock Spatial analyses of immune cell infiltration in cancer: current methods and future directions: A report of the international immuno-oncology biomarker working group on breast cancer.
\newblock \emph{The Journal of pathology}, 260\penalty0 (5):\penalty0 514--532, 2023.

\bibitem[Qi et~al.(2017{\natexlab{a}})Qi, Su, Mo, and Guibas]{qi2017pointnet}
Charles~R Qi, Hao Su, Kaichun Mo, and Leonidas~J Guibas.
\newblock Pointnet: Deep learning on point sets for 3d classification and segmentation.
\newblock In \emph{Proceedings of the IEEE conference on computer vision and pattern recognition}, pages 652--660, 2017{\natexlab{a}}.

\bibitem[Qi et~al.(2017{\natexlab{b}})Qi, Yi, Su, and Guibas]{qi2017pointnetpp}
Charles~Ruizhongtai Qi, Li Yi, Hao Su, and Leonidas~J Guibas.
\newblock Pointnet++: Deep hierarchical feature learning on point sets in a metric space.
\newblock \emph{Advances in neural information processing systems}, 30, 2017{\natexlab{b}}.

\bibitem[Qiu et~al.(2022)Qiu, Yang, Wang, and Fu]{qiu2022ivt}
Zhongwei Qiu, Qiansheng Yang, Jian Wang, and Dongmei Fu.
\newblock Ivt: An end-to-end instance-guided video transformer for 3d pose estimation.
\newblock In \emph{Proceedings of the 30th ACM International Conference on Multimedia}, pages 6174--6182, 2022.

\bibitem[Qiu et~al.(2023)Qiu, Yang, Fu, Liu, Xu, and Fu]{qiu2023learning}
Zhongwei Qiu, Huan Yang, Jianlong Fu, Daochang Liu, Chang Xu, and Dongmei Fu.
\newblock Learning degradation-robust spatiotemporal frequency-transformer for video super-resolution.
\newblock \emph{IEEE Transactions on Pattern Analysis and Machine Intelligence}, 2023.

\bibitem[Qiu et~al.(2024)Qiu, Chao, Liu, Shen, Lu, Yan, Jin, Bian, and Jiang]{qiu2024end}
Zhongwei Qiu, Hanqing Chao, Wenbin Liu, Yixuan Shen, Le Lu, Ke Yan, Dakai Jin, Yun Bian, and Hui Jiang.
\newblock End-to-end multi-source visual prompt tuning for survival analysis in whole slide images.
\newblock \emph{arXiv preprint arXiv:2409.03804}, 2024.

\bibitem[Roetzer-Pejrimovsky et~al.(2022)Roetzer-Pejrimovsky, Moser, Atli, Vogel, Mercea, Prihoda, Gelpi, Haberler, H{\"o}ftberger, Hainfellner, et~al.]{roetzer2022digital}
Thomas Roetzer-Pejrimovsky, Anna-Christina Moser, Baran Atli, Clemens~Christian Vogel, Petra~A Mercea, Romana Prihoda, Ellen Gelpi, Christine Haberler, Romana H{\"o}ftberger, Johannes~A Hainfellner, et~al.
\newblock The digital brain tumour atlas, an open histopathology resource.
\newblock \emph{Scientific Data}, 9\penalty0 (1):\penalty0 55, 2022.

\bibitem[Saltz et~al.(2018)Saltz, Gupta, Hou, Kurc, Singh, Nguyen, Samaras, Shroyer, Zhao, Batiste, et~al.]{saltz2018spatial}
Joel Saltz, Rajarsi Gupta, Le Hou, Tahsin Kurc, Pankaj Singh, Vu Nguyen, Dimitris Samaras, Kenneth~R Shroyer, Tianhao Zhao, Rebecca Batiste, et~al.
\newblock Spatial organization and molecular correlation of tumor-infiltrating lymphocytes using deep learning on pathology images.
\newblock \emph{Cell reports}, 23\penalty0 (1):\penalty0 181--193, 2018.

\bibitem[Samuel and Hudson(2012)]{samuel2012molecular}
Nardin Samuel and Thomas~J Hudson.
\newblock The molecular and cellular heterogeneity of pancreatic ductal adenocarcinoma.
\newblock \emph{Nature reviews Gastroenterology \& hepatology}, 9\penalty0 (2):\penalty0 77--87, 2012.

\bibitem[Shao et~al.(2024)Shao, Shi, Zhang, Zhou, and Wan]{shao2024tumor}
Wei Shao, YangYang Shi, Daoqiang Zhang, JunJie Zhou, and Peng Wan.
\newblock Tumor micro-environment interactions guided graph learning for survival analysis of human cancers from whole-slide pathological images.
\newblock In \emph{Proceedings of the IEEE/CVF Conference on Computer Vision and Pattern Recognition}, pages 11694--11703, 2024.

\bibitem[Shao et~al.(2021)Shao, Bian, Chen, Wang, Zhang, Ji, et~al.]{shao2021transmil}
Zhuchen Shao, Hao Bian, Yang Chen, Yifeng Wang, Jian Zhang, Xiangyang Ji, et~al.
\newblock Transmil: Transformer based correlated multiple instance learning for whole slide image classification.
\newblock \emph{Advances in neural information processing systems}, 34:\penalty0 2136--2147, 2021.

\bibitem[Shui et~al.(2024)Shui, Zheng, Zhu, Zhang, Yu, Li, Li, Chen, and Yang]{shui2024dpa}
Zhongyi Shui, Sunyi Zheng, Chenglu Zhu, Shichuan Zhang, Xiaoxuan Yu, Honglin Li, Jingxiong Li, Pingyi Chen, and Lin Yang.
\newblock Dpa-p2pnet: Deformable proposal-aware p2pnet for accurate point-based cell detection.
\newblock In \emph{Proceedings of the AAAI Conference on Artificial Intelligence}, pages 4864--4872, 2024.

\bibitem[Song et~al.(2024)Song, Chen, Ding, Williamson, Jaume, and Mahmood]{song2024morphological}
Andrew~H Song, Richard~J Chen, Tong Ding, Drew~FK Williamson, Guillaume Jaume, and Faisal Mahmood.
\newblock Morphological prototyping for unsupervised slide representation learning in computational pathology.
\newblock In \emph{Proceedings of the IEEE/CVF Conference on Computer Vision and Pattern Recognition}, pages 11566--11578, 2024.

\bibitem[Sorbellini et~al.(2005)Sorbellini, Kattan, Snyder, Reuter, Motzer, Goetzl, McKIERNAN, and Russo]{sorbellini2005postoperative}
Maximiliano Sorbellini, Michael~W Kattan, Mark~E Snyder, Victor Reuter, Robert Motzer, Manlio Goetzl, JAMES McKIERNAN, and Paul Russo.
\newblock A postoperative prognostic nomogram predicting recurrence for patients with conventional clear cell renal cell carcinoma.
\newblock \emph{The Journal of urology}, 173\penalty0 (1):\penalty0 48--51, 2005.

\bibitem[Sun et~al.(2023)Sun, Zhu, Zheng, Zhang, Shui, Yu, Zhao, Li, Zhang, Zhao, Lyu, and Yang]{sun2023pathasst}
Yuxuan Sun, Chenglu Zhu, Sunyi Zheng, Kai Zhang, Zhongyi Shui, Xiaoxuan Yu, Yizhi Zhao, Honglin Li, Yunlong Zhang, Ruojia Zhao, Xinheng Lyu, and Lin Yang.
\newblock Pathasst: Redefining pathology through generative foundation ai assistant for pathology, 2023.

\bibitem[Vu et~al.(2019)Vu, Graham, Kurc, To, Shaban, Qaiser, Koohbanani, Khurram, Kalpathy-Cramer, Zhao, et~al.]{vu2019methods}
Quoc~Dang Vu, Simon Graham, Tahsin Kurc, Minh Nguyen~Nhat To, Muhammad Shaban, Talha Qaiser, Navid~Alemi Koohbanani, Syed~Ali Khurram, Jayashree Kalpathy-Cramer, Tianhao Zhao, et~al.
\newblock Methods for segmentation and classification of digital microscopy tissue images.
\newblock \emph{Frontiers in bioengineering and biotechnology}, 7:\penalty0 433738, 2019.

\bibitem[Wang(2023)]{wang2023octformer}
Peng-Shuai Wang.
\newblock Octformer: Octree-based transformers for 3d point clouds.
\newblock \emph{ACM Transactions on Graphics (TOG)}, 42\penalty0 (4):\penalty0 1--11, 2023.

\bibitem[Wang et~al.(2023)Wang, Rong, Zhou, Yang, Zhang, Zhan, Bishop, Chi, Wilhelm, Zhang, et~al.]{wang2023deep}
Shidan Wang, Ruichen Rong, Qin Zhou, Donghan~M Yang, Xinyi Zhang, Xiaowei Zhan, Justin Bishop, Zhikai Chi, Clare~J Wilhelm, Siyuan Zhang, et~al.
\newblock Deep learning of cell spatial organizations identifies clinically relevant insights in tissue images.
\newblock \emph{Nature communications}, 14\penalty0 (1):\penalty0 7872, 2023.

\bibitem[Wang et~al.(2024)Wang, Zhao, Marostica, Yuan, Jin, Zhang, Li, Tang, Wang, Li, et~al.]{chief}
Xiyue Wang, Junhan Zhao, Eliana Marostica, Wei Yuan, Jietian Jin, Jiayu Zhang, Ruijiang Li, Hongping Tang, Kanran Wang, Yu Li, et~al.
\newblock A pathology foundation model for cancer diagnosis and prognosis prediction.
\newblock \emph{Nature}, pages 1--9, 2024.

\bibitem[Wu et~al.(2022)Wu, Lao, Jiang, Liu, and Zhao]{wu2022point}
Xiaoyang Wu, Yixing Lao, Li Jiang, Xihui Liu, and Hengshuang Zhao.
\newblock Point transformer v2: Grouped vector attention and partition-based pooling.
\newblock \emph{Advances in Neural Information Processing Systems}, 35:\penalty0 33330--33342, 2022.

\bibitem[Wu et~al.(2024)Wu, Jiang, Wang, Liu, Liu, Qiao, Ouyang, He, and Zhao]{wu2024point}
Xiaoyang Wu, Li Jiang, Peng-Shuai Wang, Zhijian Liu, Xihui Liu, Yu Qiao, Wanli Ouyang, Tong He, and Hengshuang Zhao.
\newblock Point transformer v3: Simpler faster stronger.
\newblock In \emph{Proceedings of the IEEE/CVF Conference on Computer Vision and Pattern Recognition}, pages 4840--4851, 2024.

\bibitem[Xu et~al.(2024)Xu, Usuyama, Bagga, Zhang, Rao, Naumann, Wong, Gero, Gonz{\'a}lez, Gu, et~al.]{xu2024whole}
Hanwen Xu, Naoto Usuyama, Jaspreet Bagga, Sheng Zhang, Rajesh Rao, Tristan Naumann, Cliff Wong, Zelalem Gero, Javier Gonz{\'a}lez, Yu Gu, et~al.
\newblock A whole-slide foundation model for digital pathology from real-world data.
\newblock \emph{Nature}, pages 1--8, 2024.

\bibitem[Zhao et~al.(2019)Zhao, Jiang, Fu, and Jia]{zhao2019pointweb}
Hengshuang Zhao, Li Jiang, Chi-Wing Fu, and Jiaya Jia.
\newblock Pointweb: Enhancing local neighborhood features for point cloud processing.
\newblock In \emph{Proceedings of the IEEE/CVF conference on computer vision and pattern recognition}, pages 5565--5573, 2019.

\bibitem[Zhao et~al.(2021)Zhao, Jiang, Jia, Torr, and Koltun]{zhao2021point}
Hengshuang Zhao, Li Jiang, Jiaya Jia, Philip~HS Torr, and Vladlen Koltun.
\newblock Point transformer.
\newblock In \emph{Proceedings of the IEEE/CVF international conference on computer vision}, pages 16259--16268, 2021.

\end{thebibliography}
